\documentclass[final,5p,times,twocolumn]{elsarticle}

\usepackage{amssymb}
\usepackage{amsmath}

\usepackage{array}
\usepackage{graphicx}
\usepackage{booktabs}
\usepackage{url}
\usepackage{siunitx}
\usepackage{enumitem}
\usepackage{etoolbox}
\usepackage{makecell}
\usepackage{subcaption}
\usepackage{placeins}
\usepackage{float}
\usepackage{textcomp}
\usepackage[table]{xcolor}
\usepackage{comment}
\usepackage{algorithm}
\usepackage{algpseudocode}
\usepackage[normalem]{ulem}
\usepackage{multirow}
\usepackage{caption}
\usepackage{microtype}
\usepackage[pagebackref=false,breaklinks=true,colorlinks=true,bookmarks=false,pageanchor=false]{hyperref}

\definecolor{darkgreen}{rgb}{0.0, 0.5, 0.0}
\definecolor{robo_blue}{RGB}{66, 133, 244}
\definecolor{robo_red}{RGB}{231, 66, 52}
\definecolor{robo_yellow}{RGB}{251, 189, 5}
\definecolor{robo_green}{RGB}{51, 168, 82}
\definecolor{robo_gray}{RGB}{165, 165, 165}

\newcommand{\best}[1]{\textbf{#1}}
\newcommand{\secondbest}[1]{\underline{#1}}

\newcommand{\revtwo}[1]{#1}
\newcommand{\revthree}[1]{\textcolor{black}{#1}}

\newcommand{\SR}[1]{}

\colorlet{MScolor}{black}
\newcommand{\MS}[1]{\textcolor{MScolor}{#1}}
\newcommand{\SK}[1]{}

\hypersetup{
  colorlinks=true,
  linkcolor=blue,
  filecolor=magenta,
  urlcolor=cyan,
  citecolor=darkgreen,
  pdftitle={CAT-GS},
}

\biboptions{square,numbers,sort&compress}

\journal{}

\begin{document}
\raggedbottom
\emergencystretch=3em
\hfuzz=5pt
\hbadness=10000
\vbadness=10000
\begin{frontmatter}


  \title{CAT-GS: Balanced Multimodal Learning via Calibrated Gating and Fusion Surgery} 


\author[nsu]{Mahir Shahriar Tamim\corref{cor1}}
\ead{mahir.tamim@northsouth.edu}
\cortext[cor1]{Corresponding author}

\author[nsu]{Sharjil Khan}
\author[nsu]{Md. Samiul Alim}
\author[nsu]{Tanvir Ahmed Khan}
\author[nsu]{Shafin Rahman}
\author[nsu]{Nabeel Mohammed}

\affiliation[nsu]{organization={Department of Electrical and Computer Engineering, North South University},
            city={Dhaka},
            country={Bangladesh}}

\begin{abstract}
End-to-end training of multimodal neural networks often exhibits unstable neural dynamics characterized by three coupled failure modes that degrade learning: (i) \emph{modality imbalance}, where one branch dominates gradient-based optimization; (ii) \emph{unstable gating}, where noisy confidence cues induce erratic modality selection; and (iii) \emph{fusion interference}, where modality-specific gradients conflict at the shared fusion layer.
We propose \textbf{CAT-GS} (\textit{Calibrated, Adaptive, Thresholded Gating with Fusion Surgery}), a neural dynamics-based optimization controller for intelligent computing applications. CAT-GS operates during backpropagation without modifying model architectures, fusion modules, or task losses. Through calibration of teacher-derived reliability via temperature scaling and EMA smoothing, CAT-GS stabilizes neural dynamics using a margin-thresholded policy to switch between warm-up dropout, weak-modality prioritization, and weak-biased blending, stabilizes gradient magnitudes under aggressive gating via capped gradient-budget renormalization, and applies \emph{fusion-only} PCGrad to reduce destructive cross-modal interference at the primary shared bottleneck.
We evaluate CAT-GS on audio--visual multimodal pattern recognition benchmarks (CREMA-D, AV-MNIST, and VGGSound), a tri-modal setting (UR-FUNNY), controlled synthetic data (CG-MNIST), \MS{ and additional cross-domain benchmarks (AVE and CMU-MOSI)}. CAT-GS improves or matches fused multimodal accuracy against strong imbalance-aware baselines (including OGM-GE, G$^2$D, and UMT) across settings, and yields smoother gating behavior with fewer conflicting fusion gradients.
\end{abstract}


\begin{keyword}
Neural dynamics \sep Multimodal learning \sep Intelligent computing \sep Gradient-based optimization \sep Learning dynamics \sep Knowledge distillation \sep Calibration \sep Pattern recognition



\end{keyword}

\end{frontmatter}
\hypersetup{pageanchor=true}
\flushbottom

\section{Introduction}
\label{sec:introduction}
\begin{figure}[t]
  \centering
  \includegraphics[width=\linewidth]{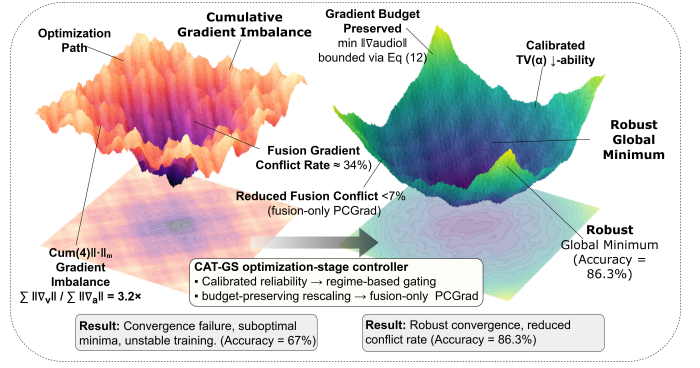} 

  \caption{\textbf{Optimization dynamics on CREMA-D.}
  Shown are experimentally derived visualizations of training behavior under standard joint training (left) and CAT-GS (right). Annotations are illustrative summaries of observed gradient statistics. Joint training exhibits unstable dynamics with dominant-modality imbalance and frequent fusion-gradient conflicts, converging to a suboptimal solution ($\approx$67\% accuracy). CAT-GS stabilizes optimization via calibrated gating, gradient-budget preservation, and fusion-only gradient surgery, yielding smooth convergence to a robust solution (86.3\% accuracy).}
  \label{fig:optimization_landscape}
\end{figure}

Deep multimodal learning integrates heterogeneous signals (e.g., audio/speech and vision) within neural network architectures to improve pattern recognition and robustness beyond any single modality \cite{Xu2023MultimodalTransformers}. From a neural dynamics perspective, intelligent computing systems require stable temporal evolution of gradient flows and adaptive control mechanisms throughout training.
CAT-GS is a neural dynamics-based optimization controller that operates during backpropagation to regulate learning dynamics without modifying model architectures, fusion modules, or task losses.
In an ideal system, the model dynamically relies on the most reliable modality \emph{for the current sample and training stage}.
In practice, however, joint optimization is frequently derailed by the \emph{modality imbalance phenomenon} \cite{ogmge2023}: one branch (often the easier or higher-SNR modality) quickly dominates gradient flow, while the weaker branch receives progressively smaller updates.
The result is brittle fusion, under-trained modality-specific encoders, and reduced multimodal gains.
Figure~\ref{fig:optimization_landscape} illustrates these failure modes on CREMA-D, highlighting how dominant-modality imbalance and fusion-gradient conflicts can derail convergence under standard joint training.

Existing approaches mitigate imbalance primarily through two families of techniques.
\emph{Gradient-modulation methods} attenuate the dominant modality’s gradients to amplify weaker ones \cite{ogmge2023, mslr2023}, while \emph{confidence-based gating methods} down-weight or mask unreliable modalities based on calibration-aware confidence estimates \cite{Han2022DynamicalFusion}.
Although effective in isolation, these strategies fail to address three tightly-coupled challenges that arise in the learning dynamics of end-to-end multimodal optimization.
First, confidence signals used for modality gating are typically derived from raw logits, which are frequently miscalibrated and noisy \cite{Guo2021OnCalibration}.
Without temporal stabilization, these unreliable estimates induce sharp batch-to-batch fluctuations in gating decisions, leading to ``gate thrashing'' rather than reliability-aware control.
Second, aggressive or persistent suppression of a modality can result in \emph{gradient starvation}: when a branch receives near-zero gradients over extended periods, its representations stagnate relative to the dominant modality, reinforcing the imbalance and making recovery increasingly difficult \cite{Rakib2025G2D}.
Third, even when modality-specific gradients are individually balanced, their interaction in the shared fusion layer can remain destructive; gradients with negative cosine similarity \cite{Yu2020GradientSurgery} interfere at the bottleneck, impeding joint convergence despite otherwise well-behaved unimodal updates.

To address these limitations, we introduce \textbf{CAT-GS}, a neural dynamics-based optimization framework for intelligent computing applications that provides a unified treatment of these coupled failure modes for calibrated, balanced, and conflict-aware multimodal learning.
CAT-GS wraps an existing multimodal training loop and modifies \emph{only} the backward-stage optimization signals.
It uses a \emph{margin-thresholded controller} driven by stabilized teacher reliability scores to switch between warm-up dropout, dominance suppression, and weak-biased blending.
This produces interpretable regime transitions and avoids oscillatory gating behavior.
Empirically, CAT-GS improves fused multimodal accuracy in diverse settings, achieving gains over strong imbalance-aware baselines (e.g., in the UR-FUNNY A--V--TXT task; Table~\ref{tab:avt-results}), while also exhibiting smoother gating dynamics and fewer negative fusion-gradient cosine events (Figure~\ref{fig:conflict_dynamics}).
Unlike prior work that addresses imbalance, gating, or conflict resolution in isolation, CAT-GS couples calibrated reliability estimation via knowledge distillation, regime-aware gating, budget-preserving gradient renormalization, and localized conflict resolution within a single optimization step.

\noindent\textbf{Novelty and contributions.}
We introduce \textsc{CAT-GS}, an optimization-stage multimodal backward-pass controller that jointly addresses gate instability, modality starvation, and fusion interference within a single optimization step.
\label{r2:novelty}%
\revtwo{The building blocks draw on established techniques; the two new elements are restricting PCGrad to the fusion bottleneck rather than the whole network, and a budget-preserving renormalization that prevents a gated modality from starving.} Our contributions are as follows:
\begin{itemize}
  \item We \revtwo{cast adaptive modality gating as} a \emph{regime-based gating controller} driven by calibrated reliability margins from knowledge distillation, enabling interpretable transitions between warm-up dropout, dominance suppression, and weak-biased blending, without modifying model architectures or task losses.
  \item We \revtwo{couple gating with} a \emph{budget-preserving gradient renormalization mechanism} that stabilizes update magnitudes under aggressive (hard) gating using an EMA target magnitude with a hard cap, mitigating failure modes such as prolonged imbalance and brittle convergence.
  \item We \revtwo{adopt} a \emph{fusion-only gradient surgery strategy}, applying PCGrad exclusively at the shared fusion bottleneck where cross-modal gradients first interact, thereby reducing destructive interference while preserving unimodal encoder dynamics \revtwo{at substantially lower cost than global projection}.
  \item We provide extensive empirical validation across audio--visual, tri-modal, and controlled synthetic benchmarks, plus additional AVE and CMU-MOSI evaluations, demonstrating consistent improvements in fused accuracy, optimization stability, and robustness to modality imbalance.
\end{itemize}

\section{Related Work}
\label{sec:related_work}

\noindent\textbf{Modality Imbalance in Multimodal Learning.}
The dominance of specific modalities during joint training is a pervasive issue in multimodal learning.
Early efforts address this through auxiliary losses or feature norm regularization~\cite{ogmge2023}.
More recent gradient-modulation techniques, such as on-the-fly rescaling~\cite{ogmge2023} and multi-loss balancing~\cite{kontras2024improving}, dynamically adjust updates to prevent one modality from overpowering others.
\MS{
Recent balancing-oriented methods include MMCosine~\cite{xu2023mmcosine}, which uses cosine-based feature normalization for improved fine-grained discriminability, DRL~\cite{wei2024diagnosing}, which diagnoses modality learning states and applies soft re-learning, and MCR~\cite{kontras2024balancing}, which introduces game-theoretic regularization to mitigate modality competition.}
While these methods improve balance, many still rely on predefined or coarse dominance cues and may not explicitly prevent long-horizon gradient starvation in suppressed branches~\cite{pezeshki2021gradient}.

In contrast, CAT-GS leverages calibrated confidence margins for context-aware modulation, ensuring suppression is temporary and paired with budget-preserving rescaling.

\noindent\textbf{Confidence-Based Gating and Dynamic Fusion.}
Multimodal fusion has been studied across a wide range of tasks, from sentiment analysis that combines audio, visual, and textual cues~\cite{poria2016fusing} to video captioning via hierarchical attention-based fusion~\cite{wu2018hierarchical}.
Gating mechanisms have long enabled adaptive fusion in multimodal systems~\cite{arevalo2017gated}.
Contemporary approaches use soft attention~\cite{tsai2019multimodal} or entropy-driven gates to weigh modalities by reliability.
Teacher-guided fusion further enhances this by distilling unimodal priors~\cite{umt2024}.
However, reliance on uncalibrated teacher logits leads to unstable decisions~\cite{Guo2021OnCalibration}, exacerbating gate thrashing.
Few works integrate direct calibration into the gating loop; CAT-GS advances this by combining temperature scaling, EMA smoothing, and margin thresholding for robust, regime-aware control.

\noindent\textbf{Gradient Conflict Resolution.}
Gradient conflicts hinder multi-task and multimodal optimization~\cite{ wei2024boosting}.
Projection-based methods like PCGrad and CAGrad excise destructive components, while others normalize via variance reduction.
In multimodal settings, conflicts are acute at fusion layers~\cite{wei2024boosting}, yet prior applications are global rather than localized.
To the best of our knowledge, prior work has not systematically explored applying PCGrad selectively at the fusion head; CAT-GS adopts this localized strategy to preserve unimodal encoder dynamics while resolving bottleneck interference.

\noindent\textbf{Knowledge Distillation in Multimodal Settings.}
Distillation from unimodal teachers is standard for multimodal students~\cite{hinton2015distilling, umt2024}.
Methods like UMT~\cite{umt2024} match logits and features but treat teacher outputs as infallible, ignoring miscalibration~\cite{Guo2021OnCalibration}.
Recent calibration-aware variants focus on post-hoc adjustments, not online integration.
CAT-GS embeds temperature scaling and EMA stabilization within the distillation-guided gating process, yielding reliable signals for adaptive optimization.

In summary, prior work typically tackles imbalance, confidence-based fusion, and gradient conflicts in isolation.
\MS{CAT-GS integrates these ideas into a single optimization-stage controller, yielding a practical recipe for stable and balanced multimodal training across fusion designs.}
\label{r2:relwork}%
\revtwo{The contribution is therefore the optimization-stage integration itself, together with the fusion-localized projection and budget-preserving renormalization it introduces.}

\section{Methodology}
\label{sec:method}
\begin{figure}[!t]
  \centering
  \includegraphics[width=\linewidth]{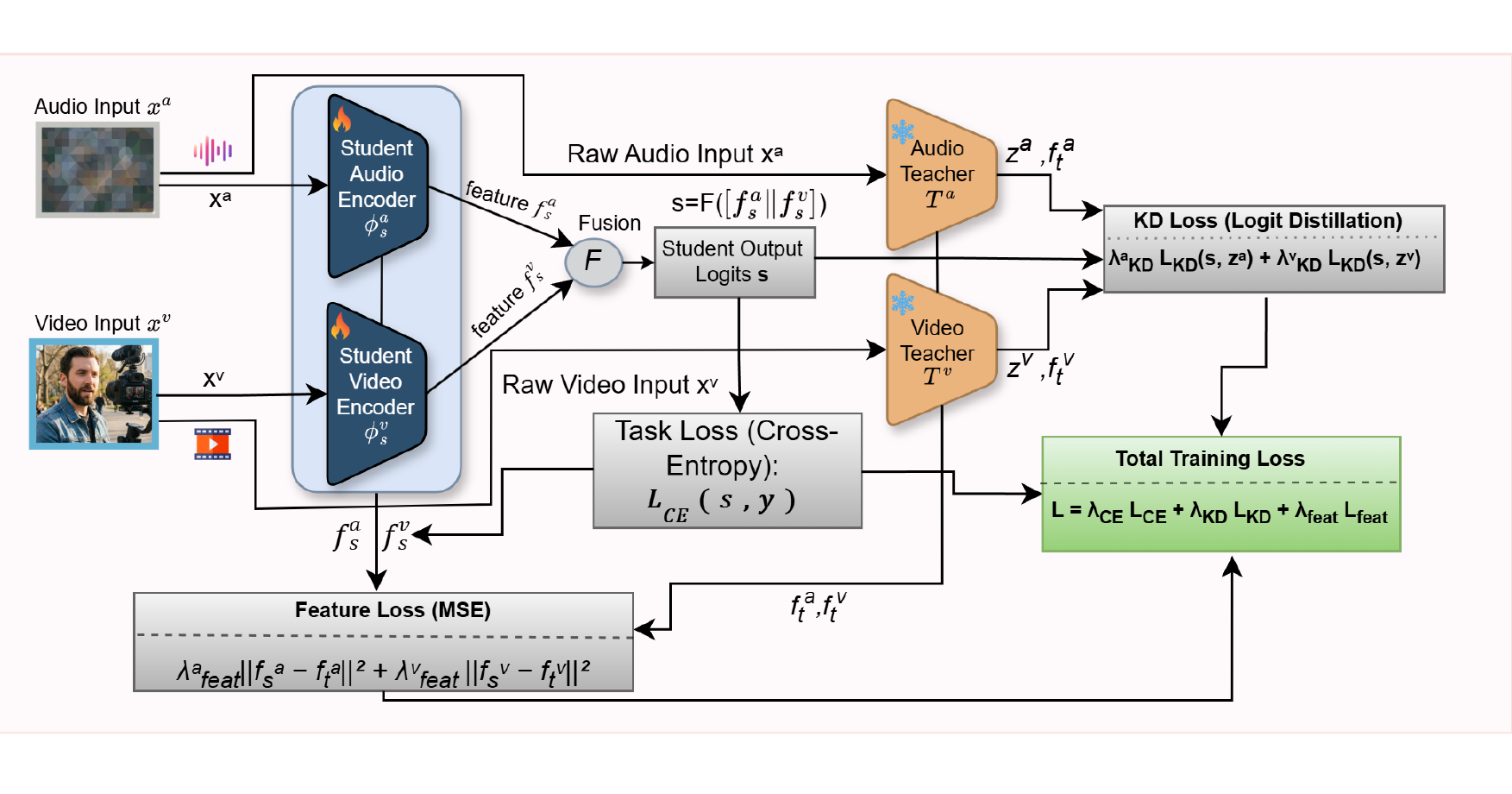} 
  \caption{
    \textbf{High-level forward architecture and supervision flow.}
    Audio and video inputs are processed by trainable student encoders and fused to produce predictions.
    \MS{Frozen unimodal teachers receive the same raw modality inputs ($x^a$, $x^v$) directly (not student-derived features) and remain frozen throughout training; they provide auxiliary supervision via knowledge distillation and per-modality reliability signals utilized by CAT-GS.}
    Optimization-stage control (CAT-GS) operates exclusively during backpropagation and is shown separately in Figure~\ref{fig:catgs_overview}.
  }
  \label{fig:forward_overview}
\end{figure}

\begin{figure*}[t]
  \centering
  \includegraphics[width=0.7\linewidth]{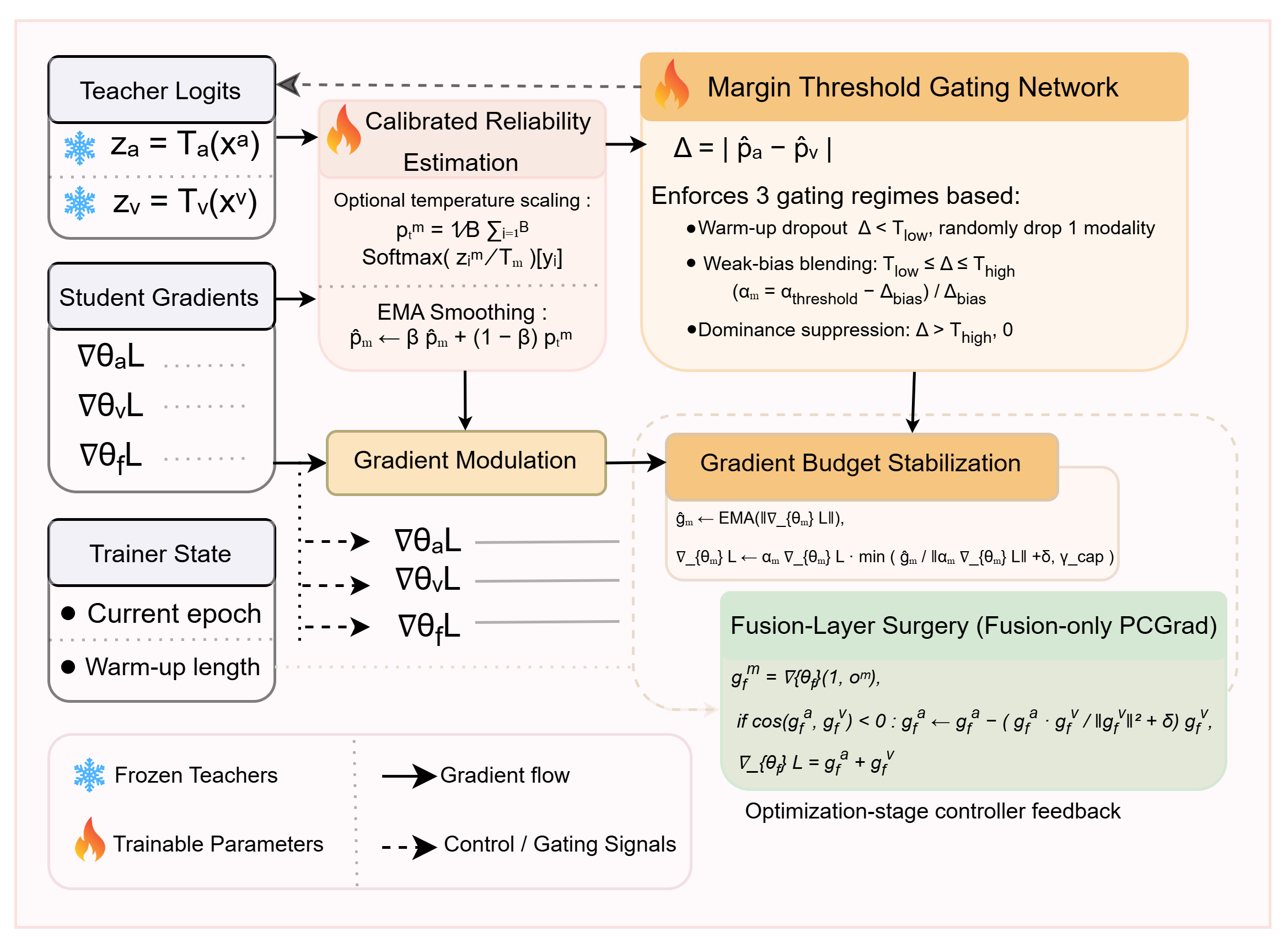}
  \caption{
    \textbf{Overview of CAT-GS.}
    The student model contains trainable unimodal encoders and a fusion head, while unimodal teachers are frozen and used only to estimate per-modality reliability.
    Teacher confidences are calibrated and stabilized to drive margin-threshold gating, followed by gradient-budget stabilization and fusion-only PCGrad.
  }
  \label{fig:catgs_overview}
\end{figure*}

Figures~\ref{fig:forward_overview} and~\ref{fig:catgs_overview} summarize the overall model and the CAT-GS training iteration.
Figure~\ref{fig:forward_overview} shows the forward architecture and supervision flow (trainable student encoders and fusion head with frozen unimodal teachers), while Figure~\ref{fig:catgs_overview} details the optimization-stage controller: calibrated/EMA-stabilized teacher reliabilities drive regime-based gating, followed by gradient-budget stabilization and fusion-only surgery before the optimizer update.
\MS{For readability, we present CAT-GS in a strict control flow: reliability estimation (Section~\ref{subsec:calibration}), regime selection and gating (Section~\ref{subsec:gating}), magnitude stabilization (Section~\ref{subsec:budget}), and fusion-conflict handling (Section~\ref{subsec:pcgrad}), followed by the unified training step in Algorithm~\ref{alg:catgs}.}

In this section we first formalize the multimodal learning setup and training objective, and then introduce CAT-GS as an optimization-level controller that operates on the resulting gradients.

\subsection{Problem Formulation}
\label{subsec:problem}

We consider a multimodal dataset
\(
\mathcal{D}=\{(x_i^a, x_i^v, y_i)\}_{i=1}^B
\)
containing paired audio--visual samples and labels.  
\MS{The student model follows a standard multimodal fusion design and comprises two unimodal encoders}
\( E_a(\cdot;\theta_a) \) and \( E_v(\cdot;\theta_v) \),
which extract modality-specific features, and a fusion network
\( F(\cdot;\theta_f) \)
that maps the concatenated representations to final logits.
Two unimodal teacher models \(T^a\) and \(T^v\) provide auxiliary guidance in the form of logits and intermediate features.
This setup is intentionally generic: the encoders may be CNNs or transformers, and the fusion head may be an MLP or attention module, since CAT-GS assumes no architectural constraints.

Given a batch, the student produces
\begin{equation}
f_s^a = E_a(x^a), \qquad
f_s^v = E_v(x^v), \qquad
\mathbf{s} = F([f_s^a \Vert f_s^v]).
\end{equation}

\noindent\textbf{Training objective.}
CAT-GS is objective-agnostic, but in our experiments we optimize a standard supervised loss augmented with unimodal-teacher distillation terms.
Let $\mathcal{M}$ denote the set of available modalities (e.g., $\{a,v\}$ or $\{a,v,t\}$). For a labeled batch, the student is trained with
\begin{equation}
\begin{aligned}
\mathcal{L}
&= \lambda_{\mathrm{CE}}\,\mathcal{L}_{\mathrm{CE}}(\mathbf{s},y)
+ \sum_{m\in\mathcal{M}} \lambda_{\mathrm{KD}}^m\,
\mathcal{L}_{\mathrm{KD}}(\mathbf{s},\mathbf{z}^m,\tau_m) \\
&\quad + \sum_{m\in\mathcal{M}} \lambda_{\mathrm{feat}}^m\,
\mathcal{L}_{\mathrm{feat}}(f_s^m,f_t^m).
\end{aligned}
\label{eq:total_loss}
\end{equation}

where $\mathcal{L}_{\mathrm{CE}}$ is cross-entropy, $f_t^m$ and $\mathbf{z}^m$ are the unimodal teacher features and logits, and
\begin{equation}
\mathcal{L}_{\mathrm{KD}}(\mathbf{s},\mathbf{z}^m,\tau_m)=\tau_m^2\,\mathrm{KL}\big(\mathrm{Softmax}(\mathbf{z}^m/\tau_m)\,\|\,\mathrm{Softmax}(\mathbf{s}/\tau_m)\big),
\end{equation}
with $\mathcal{L}_{\mathrm{feat}}$ implemented as MSE.
The coefficients $\{\lambda\}$ and temperatures $\{\tau_m\}$ match the training code and are set per experiment; when a distillation term is not used, its weight is set to zero.

Any differentiable training objective defined over \(\mathbf{s}\), the labels \(y\), and optionally the teacher outputs \(\{T^m(x^m)\}_{m\in\{a,v\}}\) induces gradients with respect to the modality-specific parameters \(\theta_a,\theta_v\) and the fusion parameters \(\theta_f\).
In a conventional setting, these gradients are aggregated and applied uniformly across modalities during backpropagation, which is precisely where modality imbalance can arise: one branch may dominate the updates, while the other receives progressively weaker signals.
CAT-GS intervenes at this optimization stage.
Rather than changing the architecture or the underlying training objective, it modulates and stabilizes modality-specific and fusion-layer gradients, aiming to ensure balanced, reliability-aware updates throughout training.

\subsection{CAT-GS Overview}
\label{subsec:catgs_overview}

Multimodal models frequently suffer from \emph{modality imbalance}, where one modality converges faster or produces cleaner gradients and consequently dominates optimization.
As training progresses, gradients for the weaker modality diminish, feature representations diverge, and the fusion layer receives conflicting update signals, ultimately leading to brittle and suboptimal fusion.
CAT-GS addresses this issue at the \emph{optimization level}.
Rather than modifying architectures or introducing task-specific losses, CAT-GS restructures how gradients are weighted, stabilized, and combined during training.
It operates transparently between \texttt{loss.backward()} and \texttt{optimizer.step()}, and can be integrated into any multimodal model with modality-specific parameters.
The framework consists of four complementary components:
\begin{enumerate}
\item \textbf{Calibrated reliability estimation} to produce stable modality confidence signals;
\item \textbf{Margin-thresholded adaptive gating} to dynamically prioritize modalities;
\item \textbf{Gradient-budget stabilization} to control update magnitudes under aggressive gating;
\item \textbf{Fusion-layer gradient surgery} to remove destructive cross-modal conflicts.
\end{enumerate}

These components are jointly necessary.
Calibration stabilizes gating decisions, adaptive gating prevents runaway dominance, budget reallocation stabilizes update magnitudes under aggressive gating, and fusion surgery resolves residual gradient interference at the bottleneck.
Together, they yield a stable optimization dynamic that adapts to evolving modality reliability without altering the forward computation.

\subsection{Calibrated Reliability Estimation}
\label{subsec:calibration}

Teacher logits encode useful task knowledge but can be noisy and overconfident, especially early in training.
Since CAT-GS relies on teacher reliability as a control signal, we stabilize these estimates to reduce batch-to-batch volatility.

\paragraph{Temperature scaling}
For the teacher $T_m$ with logits $z^m$, we compute the average probability assigned to the ground-truth label under a temperature $T_m$:
\begin{equation}
p_t^m = \frac{1}{B}\sum_{i=1}^B
\mathrm{Softmax}(z_i^m / T_m)[y_i],
\label{eq:teacher_confidence}
\end{equation}
where $T_m$ controls the sharpness of the confidence signal.
In our main experiments, we set $T_m{=}1$ for all modalities (i.e., no temperature scaling) unless explicitly stated.
We include temperature scaling as an optional extension when a calibrated temperature is available, but it is not required for CAT-GS.
\MS{Thus, temperature scaling is not a required performance driver in our default setting; it is an optional calibration switch and a stress-test factor in Section~\ref{tab:teacher_miscalibration}.}
The resulting scalar $p_t^m$ is a \emph{batch-level} (step-level) proxy for modality reliability; we use the batch mean for stability and to avoid noisy per-sample gating signals.

\paragraph{EMA smoothing}
To prevent noisy fluctuations from batch to batch, we maintain:
\begin{equation}
\hat{p}^m \leftarrow
\beta\,\hat{p}^m + (1-\beta)\,p_t^m.
\label{eq:teacher_ema}
\end{equation}
\MS{The smoothed confidences $\hat{p}^a,\hat{p}^v$ are used as the continuous control signal for gating, while temperature scaling remains an optional calibration step (default $T_m{=}1$ in our main experiments).}
This reduces ``gate thrashing'' and enables regime decisions based on reliability trends rather than instantaneous noise.
In practice, we use a high momentum (e.g., $\beta=0.9$) so that the controller reacts gradually rather than abruptly.

\subsection{Margin-Thresholded Adaptive Gating}
\label{subsec:gating}

CAT-GS uses the smoothed reliabilities to compute a \emph{reliability margin}:
\begin{equation}
\Delta = |\hat{p}^a - \hat{p}^v|.
\label{eq:delta_margin}
\end{equation}
This margin measures how far apart the modalities are in terms of calibrated confidence.
Two thresholds $\tau_{\text{low}}$ and $\tau_{\text{high}}$ define three regimes reflecting distinct optimization needs.
Crucially, these regimes determine the gating coefficients $\alpha_m$ that will later be used to scale gradients flowing through each modality.
\label{eq:alpha_gating}
\MS{For a compact view, the gating controller in this subsection can be summarized as one regime map:}
\begin{equation}
\MS{
\alpha_m(\Delta)=
\begin{cases}
\alpha_m^{\mathrm{drop}}, & \Delta<\tau_{\mathrm{low}}\ \text{and epoch}<E_w,\\
\alpha_m^{\mathrm{hard}}, & \Delta>\tau_{\mathrm{high}},\\
\alpha_m^{\mathrm{soft}}, & \tau_{\mathrm{low}}\le \Delta\le \tau_{\mathrm{high}},
\end{cases}}
\label{eq:compact_gating_map}
\end{equation}

\subsubsection*{1. Warm-up Dropout ($\Delta < \tau_{\text{low}}$, early epochs)}
\MS{Early teacher signals are noisy, so CAT-GS uses stochastic modality dropout (probability $p_{\text{drop}}$) for a short warm-up to prevent premature specialization and to strengthen unimodal representations.}

\subsubsection*{2. Dominance Suppression ($\Delta > \tau_{\text{high}}$)}
When one modality becomes clearly more reliable, CAT-GS protects the weaker one by performing hard gating:
\begin{equation}
\alpha_m =
\begin{cases}
1, & \hat{p}^m < \hat{p}^{\neg m},\\
0, & \text{otherwise}.
\end{cases}
\label{eq:alpha_hard_gating}
\end{equation}
The weaker branch receives the full update emphasis during this regime, counteracting runaway dominance.
\MS{This hard regime is used only when the margin is clearly large; otherwise the controller uses soft blending.}

\subsubsection*{3. Weak-Bias Blending ($\tau_{\text{low}} \le \Delta \le \tau_{\text{high}}$)}
\MS{When reliabilities are close, CAT-GS uses smooth weak-biased blending rather than abrupt switching:}
\begin{equation}
\tilde{\alpha}_m =
\frac{\hat{p}^m}{\hat{p}^a+\hat{p}^v},
\label{eq:alpha_soft_ratio}
\end{equation}
then boosts the weaker modality using a bias factor $\lambda_{bias}$:
\begin{equation}
\alpha_{weak} = \tilde{\alpha}_{weak}(1+\lambda_{bias}),\quad
\alpha_{strong}=\tilde{\alpha}_{strong}(1-\lambda_{bias}),
\label{eq:alpha_weak_bias}
\end{equation}
\MS{Weights are then normalized with floor $\varepsilon$ so both modalities remain active while the weaker branch receives a controlled advantage.}

\MS{Together, the three regimes define an interpretable controller that outputs $\alpha_a,\alpha_v\in[0,1]$ for gradient modulation.}

\begin{equation}
\nabla_{\theta_m} \leftarrow \alpha_m\nabla_{\theta_m}.
\label{eq:grad_modulation}
\end{equation}

\subsection{Gradient-Budget Reallocation}
\label{subsec:budget}

Directly applying $\alpha_m$ scales gradients but does not guarantee stable optimization dynamics under aggressive gating.
A fundamental risk in gating mechanisms is \emph{gradient starvation}: if a modality is consistently down-weighted (or hard-gated), its encoder can receive near-zero gradients over extended periods.
This causes a vicious cycle where the suppressed modality's features stagnate, making it even less reliable in future epochs, effectively permanently disabling a branch of the network.

CAT-GS mitigates these dynamics by (i) enforcing a nonzero floor $\varepsilon$ in soft regimes (Section~\ref{subsec:gating}), and (ii) stabilizing gradient magnitudes during hard-gating events via \emph{gradient-budget reallocation}.
We track the historical gradient norm (per modality) as a running reference scale:
\begin{equation}
\hat{g}^m \leftarrow
\beta_g \hat{g}^m + (1-\beta_g)\|\nabla_{\theta_m}\|.
\label{eq:budget_ema}
\end{equation}
After gating, we rescale gradients:
\begin{equation}
\nabla_{\theta_m}
\leftarrow
\alpha_m \nabla_{\theta_m}\;
\min\!\left(
\frac{\hat{g}^m}{\|\alpha_m \nabla_{\theta_m}\|+\delta},\;
\gamma_{\text{cap}}
\right).
\label{eq:budget_rescale}
\end{equation}
\noindent\textbf{Optimization view:}
Gating alone makes encoder updates scale linearly with $\alpha_m$, so when $\alpha_m\approx 0$ for many steps, the expected update magnitude collapses and the modality may not recover.
CAT-GS instead uses $\hat{g}^m$ as a running reference scale and (when a modality is hard-gated) renormalizes the \emph{active} modality's encoder gradients toward its EMA magnitude, capped by $\gamma_{\text{cap}}$, to avoid brittle update collapse or instability.
Equivalently, Step~5 approximately enforces
\(\|\nabla_{\theta_m}\| \approx \min(\hat{g}^m,\, \gamma_{\text{cap}}\,\|\tilde{\nabla}_{\theta_m}\|)\),
which stabilizes the magnitude of the applied update while avoiding exploding rescaling when the current gradient is extremely small.

This stabilizes the overall optimization dynamics under aggressive gating. When a modality is fully gated ($\alpha_m=0$), its encoder receives no update by design; CAT-GS focuses on keeping the remaining active branch well-conditioned while soft regimes maintain nonzero updates via the $\varepsilon$ floor.
The cap $\gamma_{\text{cap}}$ prevents excessively large rescaling when the current gradient is extremely small, and $\delta$ avoids numerical instability.

\subsection{Fusion-Layer Gradient Surgery}
\label{subsec:pcgrad}

Even when unimodal gradients are balanced, fusion parameters often receive \emph{conflicting} update directions.
Audio may encourage temporal invariances while visual gradients push toward spatial distinctions, producing destructive interference.
While methods like PCGrad \cite{Yu2020GradientSurgery} address this globally, applying projection to the entire network is computationally expensive ($O(N^2)$ in the number of tasks/modalities) and may interfere with the specialized feature extraction of unimodal encoders.

We identify the fusion output layer as the critical \emph{informational bottleneck} where these semantic conflicts are most destructive.
Therefore, CAT-GS applies a surgical, localized PCGrad-like projection only to the fusion output parameters (in our implementation, the final fusion classifier layer, e.g., \texttt{fc\_out}).

Let $\theta_f$ denote the parameters of the fusion output layer, and let
$\mathbf{o}^a$ and $\mathbf{o}^v$ be the modality-specific fusion outputs (logits)
obtained by passing only modality-$m$ features through the fusion head.
In the current implementation, we form modality-specific fusion gradients via
a Jacobian--vector product using an all-ones vector (equivalently, the gradient
of the sum of logits):
\begin{gather}
\mathbf{g}_f^m = \nabla_{\theta_f}\,\langle \mathbf{1}, \mathbf{o}^m\rangle, \qquad m\in\{a,v\}. \label{eq:fusion_jvp_grad} \\
\text{If } \quad \cos(\mathbf{g}_f^a,\mathbf{g}_f^v) < 0, \label{eq:pcgrad_conflict}
\end{gather}
the gradients point in conflicting directions.
CAT-GS applies a PCGrad-style projection (single-sided in the current implementation):
\begin{equation}
\mathbf{g}_f^a
\leftarrow
\mathbf{g}_f^a
-
\frac{\mathbf{g}_f^a\cdot\mathbf{g}_f^v}
{\|\mathbf{g}_f^v\|^2+\delta}\,\mathbf{g}_f^v,
\label{eq:pcgrad_projection}
\end{equation}
The filtered gradients are then summed to form the fusion-layer update, and the fusion layer's gradient is overwritten with $\mathbf{g}_f^a+\mathbf{g}_f^v$.
This approach resolves the "trilemma" component of gradient interference efficiently, ensuring that the shared decision boundary respects the geometric requirements of both modalities without the overhead of global gradient projection.

\noindent\textbf{Why fusion-only PCGrad.} Gradient conflicts in multimodal networks arise predominantly at shared representational bottlenecks, where modality-specific signals compete to shape a common decision boundary. Applying projection-based methods such as PCGrad globally treats unimodal encoders as competing tasks, which can inadvertently suppress modality-specific feature learning and increase computational overhead. CAT-GS instead applies PCGrad-like projection only at the fusion output layer, where cross-modal gradients first interact. This localized projection preserves unimodal encoder gradients while explicitly enforcing cooperative geometry at the shared fusion parameters, yielding stable convergence at substantially lower cost than full-network gradient projection.

\noindent\textbf{Cost perspective.} For $M$ modalities, projection-based conflict handling is pairwise ($O(M^2)$). Applying it globally scales with the full parameter count $|\theta|$, whereas fusion-only surgery scales with the fusion head size $|\theta_f|$ (typically $|\theta_f|\ll|\theta|$):
\(O(M^2|\theta|)\) vs. \(O(M^2|\theta_f|)\).
This substantially reduces overhead and avoids unnecessarily constraining unimodal encoder learning.

\begin{algorithm}[t]
\caption{CAT-GS Training Step}
\label{alg:catgs}
\begin{algorithmic}[1]
\State \textbf{Input:} $(x^a, x^v, y)$, teachers $T^a, T^v$, student $S$
\State \textbf{Hyperparameters:} $\tau_{\text{low}}, \tau_{\text{high}}, E_w, p_{\text{drop}}, \beta, \beta_g,$
\Statex \hspace{\algorithmicindent}$\lambda_{bias}, \gamma_{\text{cap}}, \varepsilon, \delta$

\State \textbf{1. Teacher Reliability}
\State $z^a \gets T^a(x^a)$,\; $z^v \gets T^v(x^v)$
\State Compute calibrated confidences $p_t^a, p_t^v$ (Eq.~\eqref{eq:teacher_confidence})
\State Update EMA reliabilities $\hat{p}^a, \hat{p}^v$ (Eq.~\eqref{eq:teacher_ema})

\State \textbf{2. Student Forward \& Backward}
\State $f_s^a, f_s^v, \mathbf{s} \gets S(x^a, x^v)$
\State Compute $\mathcal{L}$ and gradients $\nabla_\theta$

\State \textbf{3. Gating Regime Selection}
\State $\Delta \gets |\hat{p}^a - \hat{p}^v|$ (Eq.~\eqref{eq:delta_margin})
\If{$\Delta < \tau_{\text{low}}$ and epoch $< E_w$}
  \State Apply stochastic modality dropout with probability $p_{\text{drop}}$
\ElsIf{$\Delta > \tau_{\text{high}}$}
  \State Apply dominance gating (Eq.~\eqref{eq:alpha_hard_gating})
\Else
  \State Compute weak-bias weights $\alpha_a, \alpha_v$ (Eqs.~\eqref{eq:alpha_soft_ratio}--\eqref{eq:alpha_weak_bias})
\EndIf

\State \textbf{4. Gradient Modulation}
\State $\nabla_{\theta_m} \gets \alpha_m \nabla_{\theta_m}$ (Eq.~\eqref{eq:grad_modulation})

\State \textbf{5. Budget Reallocation}
\State Update $\hat{g}^m$ and rescale gradients (Eqs.~\eqref{eq:budget_ema}--\eqref{eq:budget_rescale})

\State \textbf{6. Fusion-Layer Surgery}
\State Compute fusion-output-layer gradients $\mathbf{g}_f^a,\mathbf{g}_f^v$ via Eq.~\eqref{eq:fusion_jvp_grad}; if $\cos(\mathbf{g}_f^a,\mathbf{g}_f^v)<0$, apply PCGrad-style projection (Eqs.~\eqref{eq:pcgrad_conflict}--\eqref{eq:pcgrad_projection}) and overwrite fusion-layer gradient with $\mathbf{g}_f^a+\mathbf{g}_f^v$

\State \textbf{7. Optimizer Update}
\State $\theta \gets \theta - \eta \nabla_\theta$
\end{algorithmic}
\end{algorithm}

\begin{table*}[t]
\centering
\caption{%
Accuracy (\%) of unimodal teachers ($T^a$, $T^v$) and multimodal students on audio--visual benchmarks.
For each dataset, the ``Audio'' and ``Video'' rows present the performance of the student’s modality-specific encoders,
and the ``Multi'' row reports the fused multimodal output. The best and second-best multimodal scores are highlighted
in bold and underlined, respectively (where applicable). \MS{For CAT-GS, we report mean$\pm$std when repeated runs are available.}}
\label{tab:av-results}
\setlength{\tabcolsep}{4pt}
\resizebox{\linewidth}{!}{

\begin{tabular}{cc|ccc|ccccccccccc|c}
	\toprule
\multicolumn{2}{c|}{\textbf{Dataset}} & $T^a$ & $T^v$ &

\begin{tabular}[c]{@{}c@{}}\textbf{Joint-}\\\textbf{Train}\end{tabular} &

\begin{tabular}[c]{@{}c@{}}
	\textbf{MSES}\\
{\small\cite{mSES2022}}
\end{tabular} &

\begin{tabular}[c]{@{}c@{}}
	\textbf{MSLR}\\
{\small\cite{mslr2023}}
\end{tabular} &

\begin{tabular}[c]{@{}c@{}}
	\textbf{AGM}\\
{\small\cite{agm2023}}
\end{tabular} &

\begin{tabular}[c]{@{}c@{}}
	\textbf{PMR}\\
{\small\cite{pmr2023}}
\end{tabular} &

\begin{tabular}[c]{@{}c@{}}
	\textbf{OGM-}\\
	\textbf{GE}~{\small\cite{ogmge2023}}
\end{tabular} &

\begin{tabular}[c]{@{}c@{}}
	\textbf{MLA}\\
{\small\cite{mla2022}}
\end{tabular} &

\begin{tabular}[c]{@{}c@{}}
	\textbf{Recon}\\
	\textbf{Boost}~{\small\cite{reconboost2024}}
\end{tabular} &

\begin{tabular}[c]{@{}c@{}}
	\textbf{MM}\\
	\textbf{Pareto}~{\small\cite{mmpareto2023}}
\end{tabular} &

\begin{tabular}[c]{@{}c@{}}
	\textbf{DLMG}\\
{\small\cite{dlmg2023}}
\end{tabular} &

\begin{tabular}[c]{@{}c@{}}
	\textbf{UMT}\\
{\small\cite{umt2024}}
\end{tabular} &

\begin{tabular}[c]{@{}c@{}}
	\textbf{G\textsuperscript{2}D}\\
{\small\cite{Rakib2025G2D}}
\end{tabular} &

\begin{tabular}[c]{@{}c@{}}
	\textbf{CAT-GS}\\
	\textbf{(Ours)}
\end{tabular}

\\
\cmidrule(lr){1-17}
\multirow{3}{*}{CREMA-D}
& Audio & 61.96 & - & 59.95 & 54.86 & 54.86 & 48.58 & 49.19 & 58.60 & 59.27 & 65.46 & 57.71 & 54.37 & 61.02 & 56.45 & 55.38 \\
& Video & - & 76.48 & 27.42 & 22.57 & 26.31 & 57.85 & 23.25 & 49.06 & 64.91 & 55.24 & 65.21 & 70.89 & 25.40 & 72.72 & 75.23 \\
& Multi & - & - & 67.47 & 60.99 & 64.42 & 78.48 & 59.13 & 72.18 & 79.70 & 75.13 & 79.82 & 83.62 & 67.61 & \secondbest{85.89} & \best{\MS{86.29$\pm$0.15}} \\
\cmidrule(lr){1-17}
\multirow{3}{*}{AV-MNIST}
& Audio & 42.70 & - & 16.05 & 27.50 & 22.72 & 38.90 & 37.60 & 24.53 & 42.26 & 42.11 & 41.50 & 41.99 & 31.55 & 39.10 & 41.28 \\
& Video & - & 65.34 & 55.83 & 63.34 & 62.92 & 63.65 & 58.50 & 55.85 & 65.30 & 65.26 & 64.28 & 65.04 & 64.08 & 65.09 & 64.89 \\
& Multi & - & - & 69.77 & 70.68 & 70.62 & 72.14 & 71.82 & 71.08 & 65.32 & 72.63 & 72.47 & 72.14 & 72.33 & \secondbest{73.03} & \best{\MS{73.21$\pm$0.08}} \\
\cmidrule(lr){1-17}
\multirow{3}{*}{VGGSound}
& Audio & 43.39 & - & 39.22 & 39.57 & 39.10 & 38.15 & 26.30 & 37.96 & 37.56 & 42.44 & 42.35 & 41.54 & 42.12 & 39.43 & \MS{39.30} \\
& Video & - & 32.32 & 18.70 & 17.85 & 18.66 & 25.65 & 7.12 & 22.64 & 32.02 & 17.94 & 18.12 & 23.65 & 23.77 & 29.88 & \MS{28.84} \\
& Multi & - & - & 50.97 & 50.76 & 50.98 & 47.11 & 33.07 & 51.45 & 51.65 & 49.69 & 50.97 & 52.74 & \secondbest{53.78} & \best{53.82} & \MS{53.24$\pm$0.24} \\
\cmidrule(lr){1-17}
\end{tabular}
}
\end{table*}

\subsection{Extension to $M$ Modalities and Unified Training Pipeline}
\label{subsec:m_modalities}

CAT-GS extends to $M{>}2$ modalities by computing per-modality stabilized reliabilities $\{\hat{p}^m\}_{m\in\mathcal{M}}$ and using their spread to trigger the same regime policy.
\revtwo{For the general $M$-modality setting, we replace the two-modality reliability gap of Eq.~\eqref{eq:delta_margin} with the max--min confidence margin over all available modalities:}
\begin{equation}
\Delta_M = \max_{m\in\mathcal{M}}\hat{p}^m - \min_{m\in\mathcal{M}}\hat{p}^m,
\end{equation}
and identify $m^-{=}\arg\min_m\hat{p}^m$ and $m^+{=}\arg\max_m\hat{p}^m$.
\MS{The same CAT-GS controller logic is used across datasets; only the number of modalities and corresponding branch/group definitions are adapted.}
\revtwo{The tri-modal UR-FUNNY benchmark ($M{=}3$, with $\mathcal{M}=\{a,v,t\}$) is one instance of this general formulation; the controller itself is not dataset-specific.}
The gating coefficients $\{\alpha_m\}$ are computed by: (i) warm-up dropout (optionally keep one random modality), (ii) dominance mode if $\Delta_M>\tau_{\text{high}}$ (set $\alpha_{m^-}{=}1$, others 0), else (iii) a floored proportional split with weak bias applied to $m^-$ and $m^+$:
\begin{equation}
\alpha_m \propto
\max\!\left(
\frac{\hat{p}^m}{\sum_{j\in\mathcal{M}}\hat{p}^j},\,
\varepsilon
\right)
\cdot
\begin{cases}
1+\lambda_{\text{bias}}, & m=m^{-},\\
1-\lambda_{\text{bias}}, & m=m^{+},\\
1, & \text{otherwise},
\end{cases}
\end{equation}

followed by renormalization.
Encoder gradients are then scaled as $\nabla_{\theta_m}\leftarrow\alpha_m\nabla_{\theta_m}$ with the same budget-preserving renormalization from Section~\ref{subsec:budget} to stabilize update magnitudes under aggressive gating.

Fusion-only PCGrad generalizes by computing modality-specific fusion gradients $\{\mathbf{g}_f^m\}$ and applying standard pairwise PCGrad projections sequentially across conflicting pairs, then updating the fusion head with the summed projected gradients.

\medskip
\noindent\textbf{Unified training pipeline.}
The previous sections introduced the CAT-GS components individually.
We now summarize how these components operate together during a single
training iteration.  As shown in Algorithm~\ref{alg:catgs}, teacher
reliability estimation generates the control signal for adaptive gating,
which determines modality-wise gradient scaling.  Gradient-budget
reallocation then stabilizes the magnitude of gated gradients, and
fusion-layer surgery removes cross-modal conflicts before parameters are
updated.

This pipeline consolidates CAT-GS into a lightweight controller applied
during optimization.  It introduces no changes to the forward computation
and only modest overhead in the backward pass, making it easily
applicable to a wide range of multimodal architectures.

\paragraph{Local Stability Analysis}
\label{subsec:theory_note}

\MS{CAT-GS is an optimization-stage controller that shapes the effective update direction without altering the objective. We provide a local descent guarantee for the induced update.}

\MS{Let $\mathcal{L}(\theta)$ denote the training objective and $\mathbf{u}_t$ the CAT-GS update after gating, budget reallocation, and fusion-only surgery:}

{\color{MScolor}
\begin{equation}
\theta_{t+1}=\theta_t-\eta\mathbf{u}_t.
\label{eq:theory_update}
\end{equation}
}

\MS{Assume: (A1) $\mathcal{L}$ is $L$-smooth; (A2) $\|\mathbf{u}_t\|\le C\|\nabla\mathcal{L}(\theta_t)\|$; and (A3)}

{\color{MScolor}
\begin{equation}
\langle \nabla\mathcal{L}(\theta_t),\mathbf{u}_t\rangle
\ge \rho\,\|\nabla\mathcal{L}(\theta_t)\|\,\|\mathbf{u}_t\|,\quad \rho\in(0,1].
\label{eq:theory_alignment}
\end{equation}
}

\MS{Then}

{\color{MScolor}
\begin{equation}
\mathcal{L}(\theta_{t+1})
\le
\mathcal{L}(\theta_t)
-\eta\left(\rho-\frac{L\eta C}{2}\right)
\|\nabla\mathcal{L}(\theta_t)\|^2.
\label{eq:theory_descent}
\end{equation}
}

\MS{Hence, for $0<\eta<\frac{2\rho}{LC}$, CAT-GS guarantees one-step monotone descent. This reflects: (i) EMA smoothing stabilizes regime transitions, (ii) budget caps control $C$, and (iii) fusion-only surgery improves alignment (larger $\rho$).}
\revtwo{This is a local, one-step descent guarantee for the CAT-GS update, not a global convergence result; the broader stabilization behavior is supported empirically in Section~\ref{sec:experiments}.}

\section{Experiments}
\label{sec:experiments}

We evaluate CAT-GS on audio--visual, tri-modal, and controlled synthetic settings to answer the following questions:
(Q1) Does CAT-GS improve multimodal accuracy compared to state-of-the-art imbalance-aware training methods?
(Q2) How effectively does CAT-GS stabilize unimodal branches and mitigate modality starvation?
(Q3) Does CAT-GS reduce gradient conflicts and gate thrashing during optimization?
(Q4) How do the individual components of CAT-GS contribute to overall performance?

\subsection{Experimental Setup}
\label{sec:exp_setup}

\noindent\textbf{Datasets.}
Following prior work on multimodal imbalance and gradient modulation, we evaluate CAT-GS on a representative mix of audio--visual, tri-modal, and controlled synthetic benchmarks.
	\textbf{CREMA-D}~\cite{cremad} is an audio--visual emotion recognition dataset with 7,442 clips, 6 emotion classes, and paired speech--face recordings.
	\textbf{AV-MNIST}~\cite{avmnist} combines spoken-digit audio with PCA-compressed MNIST images for 10-class classification, and is widely used to study modality dominance under noise.
  \textbf{VGGSound}~\cite{chen2020vggsound} is a large-scale audio--visual benchmark with diverse in-the-wild categories.
	\textbf{UR-FUNNY}~\cite{urfunny} is a tri-modal (audio, visual, text) humor detection dataset with 16k labeled segments and speaker-independent splits, enabling evaluation under stronger cross-modal interactions.
	\textbf{CG-MNIST}~\cite{kim2019learning} is a controlled synthetic variant of MNIST that allows explicit manipulation of modality correlations, making it suitable for analyzing imbalance and gradient starvation effects.
\MS{\textbf{Audio-Visual Event (AVE)}~\cite{ave2019} is an additional benchmark with 4,143 videos across 28 event categories, where videos are temporally labeled with audio--visual event boundaries.}
\MS{\textbf{CMU-MOSI}~\cite{zadeh2016mosi} is an additional benchmark containing 2,199 opinion video clips, each annotated with sentiment.}
\MS{We use AVE and CMU-MOSI as additional benchmarks to evaluate generalization beyond the core benchmark set.}
\MS{All datasets in this work are established public benchmarks obtained from official online sources.}

\medskip
\noindent\textbf{Baselines.}
We compare CAT-GS with ten representative multimodal imbalance-mitigation or gradient-modulation methods:
MSES~\cite{mSES2022},
MSLR~\cite{mslr2023},
AGM~\cite{agm2023},
PMR~\cite{pmr2023},
OGM-GE~\cite{ogmge2023},
MLA~\cite{mla2022},
MM-Pareto~\cite{mmpareto2023},
ReconBoost~\cite{reconboost2024},
DLMG~\cite{dlmg2023},
and UMT~\cite{umt2024}.
These baselines span gradient-based rebalancing, multi-objective optimization, reconstruction-based regularization, and unimodal-teacher-guided training.

\medskip
\noindent\textbf{Backbone Architectures and Hyperparameters.}
For audio--visual datasets (CREMA-D and AV-MNIST), we employ \textbf{ResNet-18} encoders~\cite{resnet18} for both audio and video modalities across student and teacher models.
For the tri-modal UR-FUNNY dataset, we use lightweight \textbf{Transformer encoders}~\cite{transformer} for text, audio, and visual streams to maintain consistent modality capacity.
\noindent\textbf{Teacher Protocol.}
To ensure high-quality guidance, all teacher models are pre-trained individually on their respective unimodal datasets until convergence.
During student training, these teachers are \textbf{frozen} and used solely for inference to provide logits for the calibration module.
This setup ensures that the reliability estimates $p_t^m$ are derived from stable, converged experts rather than evolving peers.

\begin{table}[t]
\centering
\large
\caption[UR-FUNNY bi-modal and tri-modal accuracy results]{Accuracy (\%) on the UR-FUNNY dataset for bi-modal and tri-modal combinations (A--V, A--TXT, V--TXT, and A--V--TXT).
Unimodal teacher performance for audio, visual, and text are $58.67\%$, $56.84\%$, and $63.48\%$, respectively.}
\label{tab:avt-results}
\setlength{\tabcolsep}{4pt}
\resizebox{\columnwidth}{!}{%
\begin{tabular}{cc|c|ccccc|c}
	\toprule
\multicolumn{2}{c|}{\textbf{Type}}
& \begin{tabular}[c]{@{}c@{}}\textbf{Joint-}\\\textbf{Train}\end{tabular}
& \begin{tabular}[c]{@{}c@{}}\textbf{OGM-}\\\textbf{GE}~\cite{ogmge2023}\end{tabular}
& \begin{tabular}[c]{@{}c@{}}\textbf{MM}\\\textbf{Pareto}~\cite{mmpareto2023}\end{tabular}
& \begin{tabular}[c]{@{}c@{}}\textbf{Recon}\\\textbf{Boost}~\cite{reconboost2024}\end{tabular}
& \begin{tabular}[c]{@{}c@{}}
    \textbf{UMT}\\
    {\small\cite{umt2024}}
  \end{tabular}
& \begin{tabular}[c]{@{}c@{}}
    \textbf{G\textsuperscript{2}D}\\
    {\small\cite{Rakib2025G2D}}
  \end{tabular}

& \textbf{CAT-GS} \\
\midrule
\multirow{3}{*}{A-V}
& Audio   & 57.34 & 59.76 & 61.77 & 60.53 & 54.63 & 59.05 & 58.25 \\
& Visual  & 53.92 & 53.82 & 55.73 & 57.87 & 56.44 & 58.05 & 54.37 \\
& Multi   & 61.57 & 61.87 & 61.27 & 62.07 & 60.46 & \textbf{62.98} & \uline{\MS{62.12$\pm$0.18}} \\
\midrule
\multirow{3}{*}{A-TXT}
& Audio   & 50.30 & 54.12 & 58.15 & 50.18 & 55.63 & 59.86 & 59.42 \\
& Text    & 57.44 & 58.35 & 58.45 & 56.98 & 57.75 & 58.85 & 59.30 \\
& Multi   & 62.17 & 62.47 & 62.80 & 61.06 & 62.47 & \uline{63.28} & \textbf{\MS{64.49$\pm$1.05}} \\
\midrule
\multirow{3}{*}{V-TXT}
& Visual  & 49.30 & 55.33 & 56.04 & 55.41 & 56.34 & 56.34 & 56.91 \\
& Text    & 51.21 & 58.95 & 59.15 & 50.94 & 53.82 & 56.04 & 57.60 \\
& Multi   & 62.07 & 62.98 & 61.27 & 60.07 & 63.18 & \uline{63.48} & \textbf{\MS{65.25$\pm$0.97}} \\
\midrule
\multirow{4}{*}{A-V-TXT}
& Audio   & 55.03 & 50.30 & 58.05 & 51.65 & 50.70 & 59.15 & 59.60 \\
& Visual  & 54.93 & 55.73 & 56.14 & 55.26 & 54.93 & 55.94 & 55.41 \\
& Text    & 58.25 & 55.71 & 58.55 & 56.25 & 52.72 & 58.15 & 60.72 \\
& Multi   & 62.58 & 63.68 & 62.88 & 61.37 & 63.38 & \uline{65.49} & \textbf{\MS{67.55$\pm$0.84}} \\
\bottomrule
\end{tabular}
}
\end{table}

\begin{figure}[!t]
  \centering
  \includegraphics[width=\linewidth]{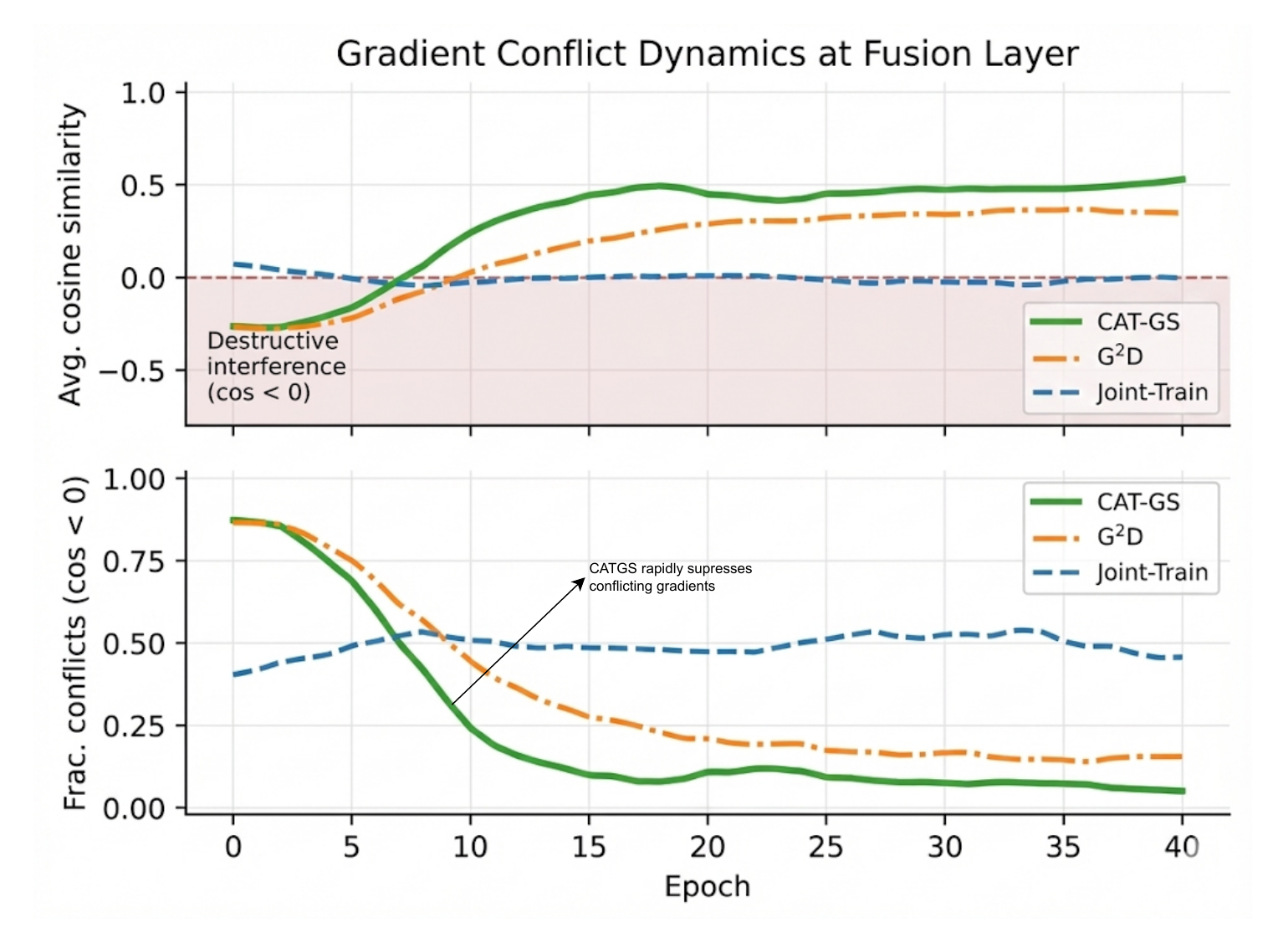}
\caption{Gradient conflict dynamics at the fusion layer. CAT-GS achieves higher gradient alignment and fewer conflicting updates than Joint-Train and G$^{2}$D, indicating effective mitigation of destructive interference.}

  \label{fig:conflict_dynamics}
\end{figure}

\begin{table}[t]
\centering
\color{MScolor}
\caption{\MS{Additional benchmark comparison on AVE and CMU-MOSI.} \revtwo{We report accuracy (\%) for representative balancing baselines and CAT-GS under the corresponding setup family} \MS{(AVE: ResNet V--A; MOSI: Transformer V--T and V--A--T).}}
\label{tab:ave-mosi-results}
\setlength{\tabcolsep}{6pt}
\renewcommand{\arraystretch}{1.08}
\resizebox{\columnwidth}{!}{%
\begin{tabular}{lccc}
\toprule
\textbf{Method} & \makecell{\textbf{AVE}\\\textbf{(ResNet V--A)}} & \makecell{\textbf{CMU-MOSI}\\\textbf{(Transformer V--T)}} & \makecell{\textbf{CMU-MOSI}\\\textbf{(Transformer V--A--T)}} \\
\midrule
MSLR~\cite{mslr2023} & 67.3 & -- & -- \\
MMCosine~\cite{xu2023mmcosine} & 65.0 & -- & -- \\
OGM~\cite{ogmge2023} & 67.3 & 73.9 & -- \\
AGM~\cite{agm2023} & 68.4 & 74.0 & 73.9 \\
ReconBoost~\cite{reconboost2024} & 68.4 & -- & -- \\
MM-Pareto~\cite{mmpareto2023} & 73.0 & 73.4 & 73.7 \\
MCR~\cite{kontras2024balancing} & 73.4 & 75.2 & 76.5 \\
DRL~\cite{wei2024diagnosing} & -- & -- & 77.99 \\
\textbf{CAT-GS (Ours)} & \textbf{74.2$\pm$0.2} & \textbf{76.0$\pm$1.2} & \textbf{78.3$\pm$1.0} \\
\bottomrule
\end{tabular}
}

\renewcommand{\arraystretch}{1.0}
\end{table}
\begin{table}[t]
\centering
\caption{Performance comparison on the CG-MNIST dataset.
Modality 1: monochromatic image; Modality 2: gray-scale image.}
\label{tab:cgmnist-results}
\setlength{\tabcolsep}{5pt}
\resizebox{\columnwidth}{!}{
\begin{tabular}{c|ccccc|c}
	\toprule
	\textbf{Modality}
& \textbf{PMR}~\cite{pmr2023}
& \textbf{OGM-GE}~\cite{ogmge2023}
& \textbf{MSES}~\cite{mSES2022}
& \textbf{MSLR}~\cite{mslr2023}
& \textbf{G\textsuperscript{2}D}~\cite{Rakib2025G2D}
& \textbf{CAT-GS} \\
\midrule
Mono  & 99.30 & 99.38 & 99.26 & 99.30 & 99.26 & 99.32 \\
Gray  & 60.40 & 67.21 & 60.85 & 59.96 & 63.36 & 67.85 \\
Multi & 78.50 & 97.35 & 93.46 & 95.04 & \uline{97.08} & \best{\MS{97.42$\pm$0.23}} \\
\bottomrule
\end{tabular}

}
\end{table}

\medskip
\noindent\textbf{Training Details.}
Unless otherwise noted, we train for 400 epochs with batch size 16 using SGD
(momentum 0.9, weight decay $10^{-4}$) and an initial learning rate of $10^{-3}$,
decayed by a factor of 0.1 at epoch 200.
We fix the CAT-GS hyperparameters $\beta = 0.9$, $\beta_g = 0.9$,
$\gamma_{\mathrm{cap}} = 1.5$, $\varepsilon = 0.1$,
$\lambda_{\mathrm{bias}} = 0.2$, warm-up epochs $E_w = 5$,
and dropout probability $p_{\mathrm{drop}} = 0.6$ across datasets,
while tuning the regime thresholds $\tau_{\mathrm{low}}$ and
$\tau_{\mathrm{high}}$ once per dataset.
\MS{In all experiments except the explicit sensitivity analysis, we use identical CAT-GS hyperparameters across datasets.}

\medskip
\noindent\textbf{Evaluation Protocol and Reproducibility.}
We report top-1 classification accuracy (\%) on the held-out test split for each benchmark.
During training, we evaluate once per epoch on a validation split and select the checkpoint with the best validation accuracy; we then report test accuracy once using this selected checkpoint.
Unless otherwise noted, we report the average test accuracy (\%) over three random seeds (42, 123, and 999).
\label{para:reviewer1.5.1}
\MS{For CAT-GS, we additionally report run-to-run variability as mean$\pm$std in Tables~\ref{tab:av-results}, \ref{tab:avt-results}, and \ref{tab:cgmnist-results}.}
\MS{With $n{=}3$ runs, 95\% confidence intervals are computed as $\bar{x} \pm t_{0.975,2}\,s/\sqrt{3}$.}
CG-MNIST experiments use small CNN encoders to isolate optimization behavior from representational capacity.
\MS{All methods use identical backbones and fusion heads to ensure fair comparison.}
Experiments are conducted on NVIDIA RTX~3060 and RTX~4050 GPUs.

\subsection{Results}
\label{sec:results}

\noindent\textbf{Audio--Visual Classification Benchmarks.}
Table~\ref{tab:av-results} reports results on CREMA-D, AV-MNIST, and VGGSound.
\MS{In Table~\ref{tab:av-results}, \textbf{Joint-Train} denotes the original model trained with standard end-to-end optimization (CAT-GS disabled) under the same backbone, fusion head, optimizer schedule, and data split as CAT-GS for fair attribution.}
Across datasets, CAT-GS consistently remains competitive with imbalance-aware baselines spanning gradient modulation (e.g., OGM-GE~\cite{ogmge2023}, AGM~\cite{agm2023}, PMR~\cite{pmr2023}) and teacher-guided training (UMT~\cite{umt2024}).

\MS{On CREMA-D, where the video modality is substantially stronger than audio, CAT-GS improves the fused prediction to 86.29\%$\pm$0.15 (vs. 85.89\% for G$^2$D), indicating that stabilizing gating and resolving fusion conflicts can yield gains beyond confidence-only suppression.}
\MS{On AV-MNIST, where modalities are closer in strength, CAT-GS remains competitive and achieves the best fused accuracy (73.21\%$\pm$0.08), suggesting the controller does not over-correct when imbalance is mild.}

\label{r2:vgg_results}%
\revtwo{On VGGSound, CAT-GS reaches 53.24\%$\pm$0.24 fused accuracy and does not surpass the strongest baselines, falling below G$^2$D (53.82\%) and UMT (53.78\%) by roughly 0.5--0.6 points. This reflects the large-scale in-the-wild regime rather than the training setup: the large label space ($\sim$300 classes) and noisy clips yield low, unstable teacher reliabilities, so the margin $\Delta$ that drives gating is weakly informative, and accuracy at this scale is dominated by encoder capacity and data volume rather than by gradient-level control. CAT-GS therefore offers limited benefit in this regime.} \MS{The unimodal branches on VGGSound follow the same pattern and remain close to G$^2$D, with slightly lower Audio/Video accuracies (39.30 vs. 39.43 and 28.84 vs. 29.88).}

\MS{For completeness, we also include a no-teacher standard-training reference on CREMA-D: training the student with only the supervised classification objective (i.e., without distillation and without CAT-GS modulation) reaches 64.78\% best fused accuracy under the same backbone, fusion setup, and training schedule, and remains substantially below CAT-GS, consistent with the relative ordering reported in Table~\ref{tab:av-results}.}

\begin{figure*}[t]
  \centering
  \includegraphics[width=\linewidth]{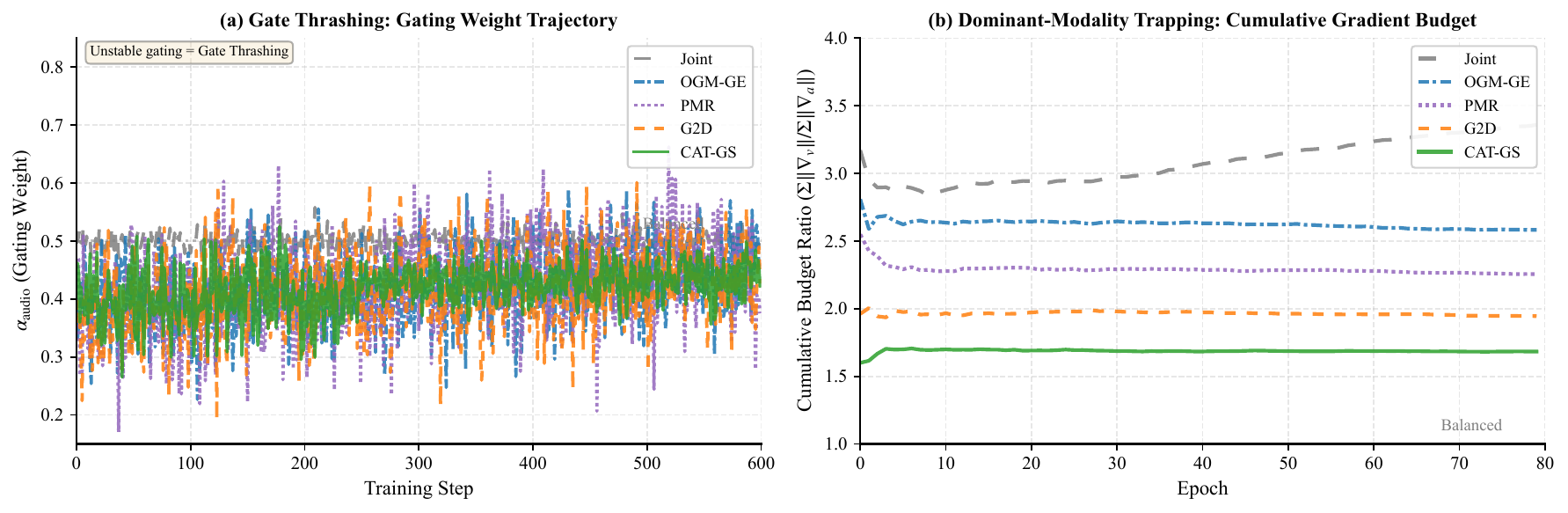}
  \caption{Optimization diagnostics on CREMA-D. (a) Audio-gating trajectory over 600 steps: OGM-GE and PMR show high-frequency oscillations (gate thrashing), whereas CAT-GS remains smooth under EMA-stabilized control. (b) Cumulative gradient-budget ratio $\Sigma\|\nabla_v\|/\Sigma\|\nabla_a\|$: Joint-Train reaches 3.2$\times$ imbalance, G$^2$D remains near 2.0$\times$, and CAT-GS reduces this to 1.7$\times$ through budget-preserving reallocation.}
  \label{fig:failure_modes}
\end{figure*}

\begin{figure}[!t]
  \centering
  \includegraphics[width=\linewidth]{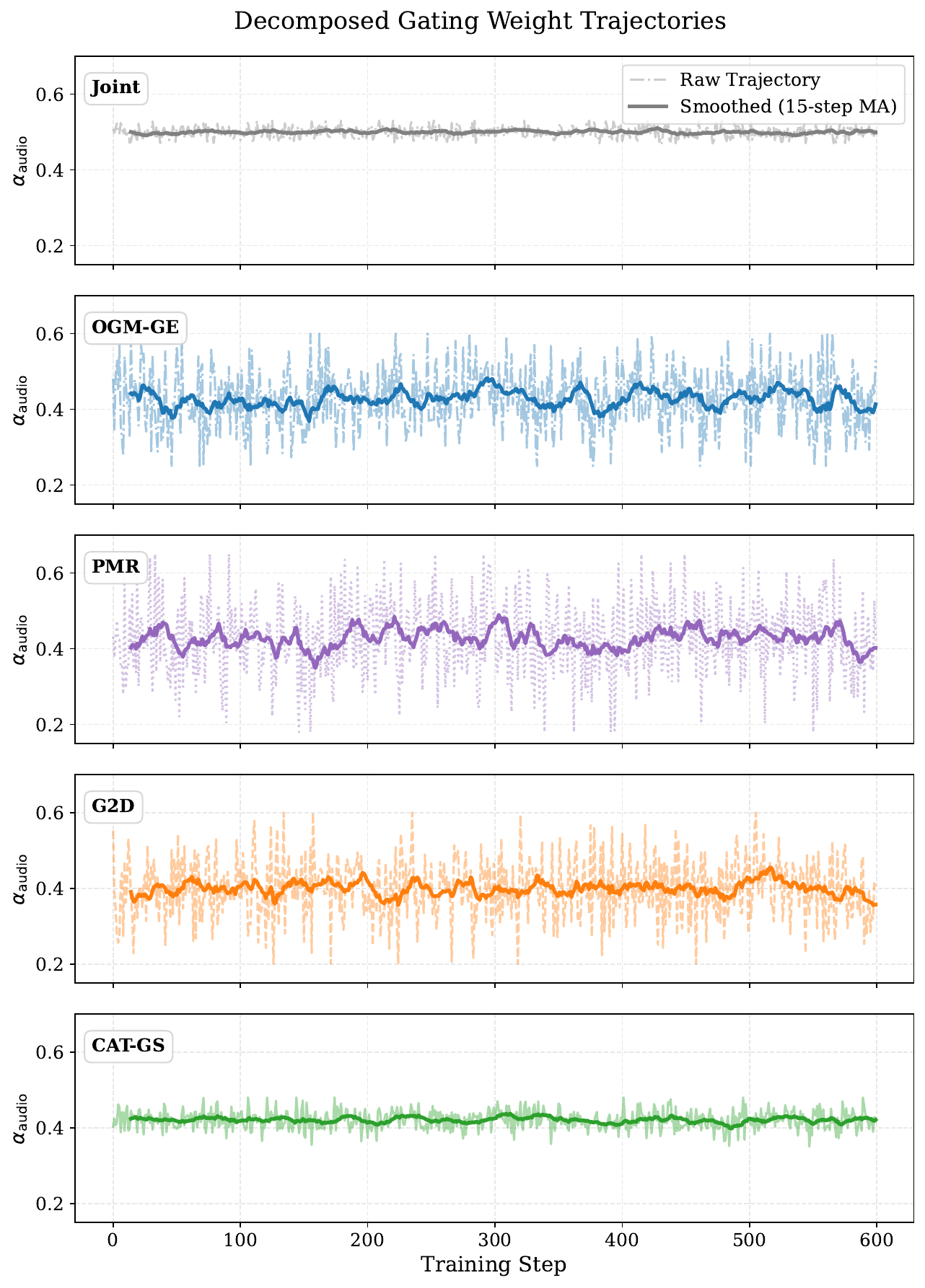}
  \caption{\MS{Decomposed gating weight trajectories ($\alpha_{\mathrm{audio}}$) across training steps for Joint, OGM-GE, PMR, G$^{2}$D, and CAT-GS.}}
  \label{fig:appendix_gating_trajectories}
\end{figure}

\begin{figure*}[!t]
  \centering
  \includegraphics[width=0.85\linewidth]{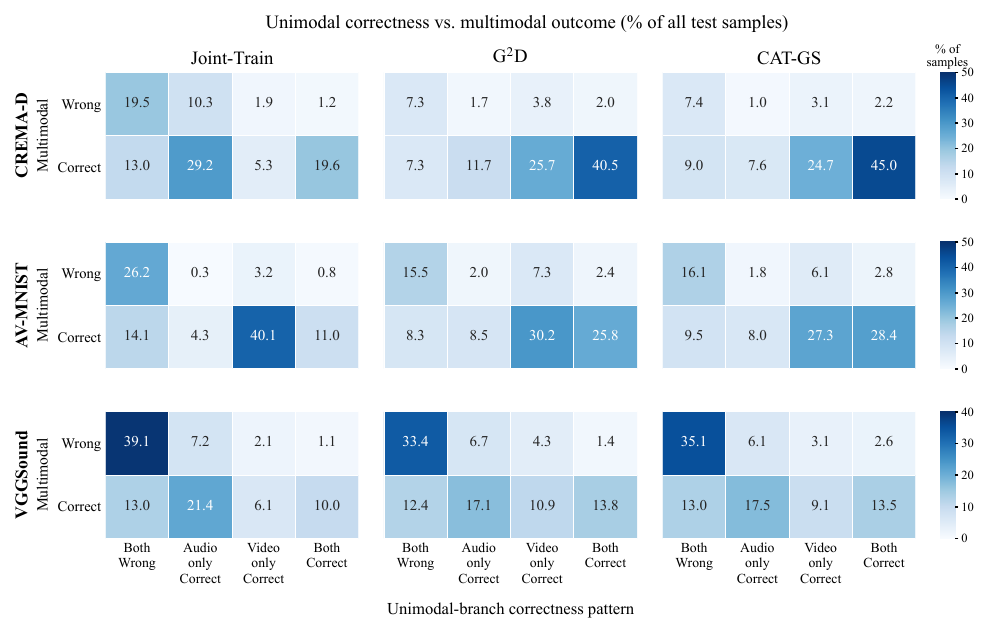}
  \caption{\MS{Unimodal correctness versus multimodal outcome for Joint-Train, G$^{2}$D, and CAT-GS (columns) on CREMA-D, AV-MNIST, and VGGSound (rows).} \revthree{Within each panel, columns group test samples by which unimodal branches are correct and rows give the multimodal outcome.}}
  \label{fig:cm_unimodal_mm}
\end{figure*}

\MS{The corresponding 95\% confidence intervals for CAT-GS are [85.92, 86.66] (CREMA-D), [73.01, 73.41] (AV-MNIST), and [52.64, 53.84] (VGGSound).}
We note that fused accuracy is the primary objective; unimodal branch accuracies can change in either direction depending on how much specialization toward fusion is beneficial.

\medskip
\noindent\textbf{Tri-Modal Humor Understanding (UR-FUNNY).}
Table~\ref{tab:avt-results} reports results on UR-FUNNY across increasing modality combinations, from audio--visual (A--V) to text-inclusive settings.
In the A--V configuration, CAT-GS performs competitively but does not outperform the strongest baseline (G\textsuperscript{2}D), which is expected given the reduced modality interactions and weaker gradient conflicts in the bi-modal setting.
In contrast, when textual modality is introduced, CAT-GS consistently achieves the best multimodal accuracy in both A--TXT and V--TXT settings, and yields the largest gains in the full A--V--TXT configuration.
Specifically, CAT-GS improves over multi-objective optimization (MM-Pareto~\cite{mmpareto2023}) and reconstruction-based regularization (ReconBoost~\cite{reconboost2024}), highlighting the effectiveness of calibrated gating, gradient-budget stabilization, and fusion-layer conflict mitigation under more challenging tri-modal fusion. \label{para:reviewer1.5.3}
\MS{For CAT-GS on UR-FUNNY, the 95\% confidence intervals are [61.67, 62.57] (A--V), [61.88, 67.10] (A--TXT), [62.84, 67.66] (V--TXT), and [65.46, 69.64] (A--V--TXT).}

\label{para:reviewer3.2}
\medskip
\noindent\textbf{\MS{Additional Benchmarks (AVE and CMU-MOSI).}}
\MS{Table~\ref{tab:ave-mosi-results} extends evaluation to additional benchmark families and includes recent baselines such as MCR and DRL. CAT-GS remains strongest on AVE and on the MOSI two-modality setting, and reaches 78.3\%$\pm$1.0 on MOSI three-modality classification, slightly above DRL (77.99) and clearly above MCR (76.5). The corresponding 95\% confidence intervals for CAT-GS are [73.70, 74.70] on AVE, [73.02, 78.98] on CMU-MOSI (V--T), and [75.82, 80.78] on CMU-MOSI (V--A--T). Compared with earlier methods (e.g., MSLR/OGM/AGM/MM-Pareto), the margins are larger, while against stronger recent methods they become tighter but remain positive. Overall, these results indicate that CAT-GS transfers to both event-centric and sentiment-centric benchmarks instead of being limited to the original dataset family.}

\medskip
\label{para:reviewer1.5.4}
\noindent\textbf{Controlled Synthetic Benchmark (CG-MNIST).}
Results on CG-MNIST are shown in Table~\ref{tab:cgmnist-results}.
This benchmark allows explicit control over label correlations between the monochromatic and grayscale modalities.
Even when one modality is made spuriously more predictive early in training, CAT-GS maintains strong fused performance and avoids brittle training dynamics observed in several baselines.
This controlled setting supports the role of margin-thresholded gating and budget-aware stabilization in reducing the risk that prolonged dominance suppresses the weaker modality.
\MS{For CAT-GS on CG-MNIST, the 95\% confidence interval is [96.85, 97.99].}

\medskip
\noindent\textbf{Evidence for Reduced Gradient Conflicts.}
To quantify cross-modal interference, we follow the implementation and define a modality-specific fusion gradient surrogate $\mathbf{g}_f^m$ on the fusion output layer (e.g., $\texttt{fc\_out}$) using the logit-sum gradients from Eq.~(\ref{eq:fusion_jvp_grad}).
We measure gradient alignment using the cosine similarity $\cos(\mathbf{g}_f^a, \mathbf{g}_f^v)$.
As shown in Figure~\ref{fig:conflict_dynamics} (top), CAT-GS maintains a positive average cosine similarity throughout training, indicating a cooperative optimization regime at the fusion layer.
Correspondingly, Figure~\ref{fig:conflict_dynamics} (bottom) shows that CAT-GS substantially reduces the fraction of training batches with negative cosine similarity, directly evidencing suppression of destructive gradient interference.

\medskip
\noindent\textbf{Gate Stability and Gradient Starvation Analysis.}
We quantify gating stability using the Total Variation (TV) of the gating coefficients.
CAT-GS demonstrates smooth gating transitions with low total variation, confirming that calibration and EMA smoothing effectively filter out high-frequency noise (``gate thrashing'').
To assess starvation risk under gating, we track the gradient norms of the unimodal encoders.
CAT-GS's $\varepsilon$-floored soft regimes reduce accidental gradient collapse, and its capped budget reallocation stabilizes update magnitudes during hard-gating events.

\medskip

\noindent\textbf{Observed failure-mode behavior.}
\label{sec:failure_modes}
Figure~\ref{fig:failure_modes} summarizes optimization dynamics from CREMA-D training logs to illustrate gate thrashing and dominant-modality trapping.
In panel (a), the audio gate $\alpha_{\mathrm{audio}}$ is plotted over 600 steps: OGM-GE and PMR exhibit pronounced oscillations under batch-wise modulation, whereas CAT-GS produces smooth trajectories using EMA-smoothed reliability ($\beta=0.9$) and margin-based regime switching.
In panel (b), we report the cumulative gradient budget ratio $\Sigma\|\nabla_v\|/\Sigma\|\nabla_a\|$.
Joint training accumulates a $3.2\times$ imbalance; gradient-modulation baselines remain around $\sim2.4\times$; G2D reaches $\sim2.0\times$; and CAT-GS maintains $1.7\times$ through budget-preserving reallocation (Eq.~\eqref{eq:budget_rescale}).
Together, these diagnostics indicate that prior approaches can suffer from gate instability and persistent gradient imbalance, which CAT-GS is designed to address.

\label{sec:decomposed_gating_trajectories}
\MS{For clearer interpretation of panel (a), Figure~\ref{fig:appendix_gating_trajectories} shows decomposed gating trajectories (raw and smoothed) for Joint, OGM-GE, PMR, G$^{2}$D, and CAT-GS. In this decomposed view, each method is shown with a raw per-step trajectory and a smoothed trend (moving average), which separates high-frequency jitter from underlying control behavior. Joint remains largely static, indicating weak reliability-driven adaptation. OGM-GE and PMR show strong high-frequency fluctuation and jagged trends, consistent with unstable batch-reactive routing. G$^{2}$D reduces fluctuation amplitude relative to OGM-GE/PMR but still exhibits frequent short-window switching. CAT-GS is the smoothest, reducing oscillation while preserving gradual trend adaptation, which is consistent with its EMA-stabilized and margin-thresholded controller design.}

\medskip
\label{para:cmexp}
\MS{To complement this optimization-dynamics view, we further provide a confusion-matrix-style qualitative analysis} \revtwo{on the three core audio--visual benchmarks (CREMA-D, AV-MNIST, and VGGSound),} \revthree{in which each test sample is assigned to a single cell according to which unimodal branches predicted it correctly and whether the fused prediction was correct (Figure~\ref{fig:cm_unimodal_mm}). The top-left cell holds the samples that neither branch predicted correctly and that fusion did not recover, which are limited by the unimodal encoders; the wrong row of the remaining columns holds those on which at least one branch was correct but the fused prediction was not, which are limited by the fusion stage. Together the two account for all errors. On CREMA-D, CAT-GS reduces both relative to Joint-Train, from $19.5\%$ to $7.4\%$ and from $13.4\%$ to $6.3\%$, consistent with the reduced negative fusion-gradient cosine events in Figure~\ref{fig:conflict_dynamics}. The fusion-stage loss is lower than that of G$^{2}$D on all three benchmarks ($6.3\%$ vs.\ $7.5\%$, $10.7\%$ vs.\ $11.7\%$, $11.8\%$ vs.\ $12.4\%$), but on VGGSound the encoder-limited error is larger ($35.1\%$ vs.\ $33.4\%$) and outweighs that advantage. This places the VGGSound shortfall in the unimodal branches rather than at the fusion layer, consistent with the weakly informative reliability margin $\Delta$ and the large label space at this scale.}

\subsection{Ablation Studies}
\label{subsec:ablation}

We perform two complementary ablations: a unified component matrix with both single-component activation and leave-one-out removal (Table~\ref{tab:catgs-unified-ablation}), and architectural ablation over fusion modules (Table~\ref{tab:fusion-results}).

\MS{Table~\ref{tab:catgs-unified-ablation} merges both ablation views using a tick-mark configuration: (i) starting from Joint-Train and enabling one component at a time, and (ii) starting from full CAT-GS and removing one component at a time. This gives a compact view of standalone contribution and component necessity in one place.}

\label{subsec:unified_ablation}
\noindent\textbf{Unified Component Ablation (Table~\ref{tab:catgs-unified-ablation}).}
\revtwo{Gating acts as the controller's hub: the calibrated, EMA-smoothed reliability margin drives only the gating regime selector, the weak-bias rule is itself a gating regime, and gradient-budget reallocation is invoked only at hard-gate ($\alpha{=}0$) events. Calibrated gating is therefore active in every variant that improves over Joint-Train, so we omit calibration and gating as columns and vary the four remaining components; removing gating reduces CAT-GS to fusion-only PCGrad, the ``+ Fusion-only PCGrad only'' row ($77.80$, $+10.33$).}
\MS{In the leave-one-out block, ``w/o Weak-Bias'' disables only the weak-bias factor by setting $\lambda_{\text{bias}}{=}0$, while keeping the adaptive gating controller (warm-up, dominance, and soft regime selection via $\tau_{\text{low}},\tau_{\text{high}}$) unchanged.}
Turning off \emph{gradient-budget reallocation} produces the largest drop in the removal block, showing that gating alone is insufficient to prevent long-term update collapse.
Discarding \emph{EMA smoothing} also causes a substantial drop, indicating strong dependence on stabilized reliability signals.
Omitting \emph{fusion-layer PCGrad surgery} hurts performance to a lesser extent, indicating conflict resolution is important but secondary to reliability and budget stabilization.
The activation block complements this view by showing each component helps over Joint-Train, while the full combination remains strongest.

\begin{table}[t]
\centering
\color{MScolor} 
\caption{\MS{Component-wise contribution and necessity analysis on CREMA-D. Checkmarks indicate enabled components.} \revtwo{Calibrated reliability and gating are the always-on base controller (not shown as columns); we vary the four components listed.}}
\label{tab:catgs-unified-ablation}
\footnotesize
\setlength{\tabcolsep}{3pt}
\resizebox{\columnwidth}{!}{%
\begin{tabular}{lcccc|cc}
\toprule
\textbf{Variant} & \textbf{EMA} & \textbf{Weak-Bias} & \textbf{Budget} & \textbf{PCGrad} & \textbf{MM} & \textbf{$\Delta_{\text{Joint}}$} \\
\midrule
\multicolumn{7}{l}{\textit{Single-component activation (from Joint-Train)}} \\
Joint-Train (baseline)              & --         & --         & --         & --         & 67.47 & 0.00 \\
\revtwo{+ Calibrated+EMA reliability} & \checkmark & --         & --         & --         & 78.90 & +11.43 \\
\revtwo{+ Adaptive gating (incl.\ weak-bias)} & --         & \checkmark & --         & --         & 77.20 & +9.73 \\
+ Gradient-budget only              & --         & --         & \checkmark & --         & 79.40 & +11.93 \\
+ Fusion-only PCGrad only           & --         & --         & --         & \checkmark & 77.80 & +10.33 \\
\midrule
\multicolumn{7}{l}{\textit{Leave-one-out removal (from full CAT-GS)}} \\
CAT-GS (full)                       & \checkmark & \checkmark & \checkmark & \checkmark & \textbf{86.29} & \textbf{+18.82} \\
w/o EMA smoothing                   & --         & \checkmark & \checkmark & \checkmark & 84.60 & +17.13 \\
w/o Weak-Bias                       & \checkmark & --         & \checkmark & \checkmark & 85.50 & +18.03 \\
w/o Grad. Budget                    & \checkmark & \checkmark & --         & \checkmark & 83.20 & +15.73 \\
w/o PCGrad                          & \checkmark & \checkmark & \checkmark & --         & 85.08 & +17.61 \\
\bottomrule
\end{tabular}%
}
\end{table}

\medskip

\begin{table}[!t]
\centering
\caption{Performance comparison of \textbf{CAT-GS} under different fusion strategies across multiple datasets in terms of accuracy (\%).}
\label{tab:fusion-results}
\resizebox{\linewidth}{!}{%
\begin{tabular}{c|cccc}
	\toprule
	\textbf{Fusion}
& \textbf{CREMA-D}
& \textbf{AV-MNIST}
& \textbf{CG-MNIST}
& \textbf{UR-FUNNY}  \\
\cmidrule(lr){1-5}

Sum
& 82.10
& 71.85
& 94.26
& 64.02 \\

Concat
& 84.72
& 72.93
& 95.88
& 65.01 \\

FiLM \cite{film}
& 83.95
& 72.12
& 92.43
& 63.77 \\

BiGated \cite{gated}
& 81.88
& 72.44
& 91.36
& 63.25 \\

Cross-Attention \cite{cross-attention}
& 85.04
& 72.98
& 96.73
& 66.41 \\

\MS{Transformer Fusion \cite{tsai2019multimodal}}
& \MS{85.73}
& \MS{72.95}
& \MS{97.06}
& \MS{66.56} \\

\MS{Early Fusion \cite{gunes2005earlylatefusion}}
& \MS{82.62}
& \MS{72.18}
& \MS{93.45}
& \MS{64.02} \\

  \textbf{Late Fusion \cite{gunes2005earlylatefusion}}
& \textbf{86.29}
& \textbf{73.21}
& \textbf{97.42}
& \textbf{67.55} \\
\bottomrule
\end{tabular}
}
\end{table}

\medskip
\noindent\textbf{\MS{Fusion Strategy Analysis} (Table~\ref{tab:fusion-results}).}
\label{para:fusion_strategy_analysis}
\MS{Table~\ref{tab:fusion-results} compares different fusion modules when used inside CAT-GS.
We explore simple additive fusion (Sum), concatenation (Concat), FiLM-style conditioning~\cite{film}, bidirectional gating (BiGated)~\cite{gated}, cross-attention~\cite{cross-attention}, and explicit Early/Late Fusion variants~\cite{gunes2005earlylatefusion,atrey2010multimodalfusion}.
We observe three trends.
First, naive Sum fusion is consistently weaker than more expressive alternatives, especially on CREMA-D and CG-MNIST, suggesting that richer cross-modal interactions remain beneficial even when gradients are well-balanced.
Second, cross-attention and concatenation form a strong pair of contenders, often approaching the best performance.}

\MS{We additionally tested a lightweight transformer-style token-fusion variant~\cite{tsai2019multimodal} under the same training protocol; CAT-GS remained stable and trainable, while absolute performance differences remained architecture-dependent.} \MS{In our experiments, Late Fusion is the primary setting (best absolute accuracy), and we report Early Fusion and Transformer Fusion as transfer checks under the same training protocol.}

The fusion ablation isolates the role of the architecture used at the multimodal head while keeping the CAT-GS controller fixed.
FiLM and BiGated fusion can already capitalize on CAT-GS, but their performance is inconsistent across datasets, likely due to the stronger inductive assumptions they impose (e.g., one modality modulating the other).
\MS{Cross-attention provides a strong trade-off and often matches or slightly trails the strongest-performing variant across datasets.}
\MS{To assess transfer across fusion heads, we evaluate Transformer Fusion, Early Fusion, and Late Fusion under the same training protocol. CAT-GS remains stable in all cases, with expected architecture-dependent absolute differences.}
\MS{Taken together, Tables~\ref{tab:catgs-unified-ablation} and~\ref{tab:fusion-results} show that (i) each component of CAT-GS contributes complementary gains, and (ii) the method is not overly sensitive to the specific fusion module.}

\medskip
\noindent Overall, CAT-GS delivers consistent gains across classification, tri-modal understanding, and controlled synthetic settings.
The ablations confirm that its advantages arise from the combination of stabilized reliability signals, adaptive regime switching, budget-aware gradient stabilization, and conflict-aware fusion.

\medskip
\noindent\textbf{Teacher Miscalibration Stress Test (Table~\ref{tab:teacher_miscalibration}).}
CAT-GS relies on unimodal teachers to provide modality reliability signals. To assess robustness to errors in these signals, we introduce a controlled teacher miscalibration stress test by applying a temperature mismatch at teacher inference time. For modality $m$, teacher logits are perturbed as $z_m \leftarrow z_m / T^{m}_{\text{mis}}$ before computing the confidence used by the CAT-GS controller.

We evaluate four regimes: (R0) a calibrated baseline, (R1) an overconfident audio teacher, (R2) an underconfident audio teacher, and (R3) both modalities miscalibrated. To isolate the role of reliability stabilization, we further evaluate the worst-performing mismatch (R2) with EMA smoothing disabled (R4) and temperature calibration removed (R5). All other settings are kept identical. For R5, we bypass the calibration step when computing reliability; to avoid confusion, we omit the mismatch-temperature entries in Table~\ref{tab:teacher_miscalibration}.
Table~\ref{tab:teacher_miscalibration} reports multimodal (MM) accuracy under each condition. CAT-GS degrades gracefully under moderate miscalibration (R1--R2), remaining close to the calibrated baseline. Performance drops further under the stressed setting (R3) but remains stable. In contrast, removing EMA smoothing (R4) or temperature calibration (R5) leads to a larger degradation, highlighting the importance of reliability stabilization. Overall, CAT-GS is robust to moderate teacher miscalibration, with stabilization mechanisms playing a critical role.
\label{para:review1.3.1}
\MS{R4--R5 are best interpreted as internal stress controls (ablation-style reliability degradation), not as a standalone weak-teacher/domain-shift benchmark. Under these stressed settings, CAT-GS remains trainable and stable, but final accuracy decreases relative to the strong-teacher setting. This supports a balanced conclusion: robustness to moderate unreliability, with clear dependence on teacher quality under stronger reliability degradation.}

\begin{table}[t]
\centering
\caption{Teacher miscalibration stress test on CREMA-D dataset. Multimodal (MM) accuracy of the fused model under temperature-mismatched teachers.}
\label{para:reviewer3.3.3}
\label{tab:teacher_miscalibration}
\resizebox{\columnwidth}{!}{%
\begin{tabular}{lcccc}
\hline
Run & $T^{a}_{\text{mis}}$ & $T^{v}_{\text{mis}}$ & MM Acc. (\%) $\uparrow$ & $\Delta$ (\%) $\downarrow$ \\
\hline
R0 (Baseline)              & 1.0 & 1.0 & 86.3 & 0.0 \\
R1 (Audio overconf.)       & 0.5 & 1.0 & 85.2 & $-1.1$ \\
R2 (Audio underconf.)      & 2.0 & 1.0 & 85.1 & $-1.2$ \\
R3 (Both mismatched)       & 2.0 & 0.5 & 84.4 & $-1.9$ \\
\hline
R4 (R2 w/o EMA)            & 2.0 & 1.0 & 83.6 & $-2.7$ \\
R5 (R2 w/o calibration)    & \MS{--} &  \MS{--} & 84.0 & $-2.3$ \\
\hline
\end{tabular}}
\end{table}

\label{para:mod_imb}
\medskip
\noindent\textbf{Modality-Imbalance Validation (clean vs degraded modality).}
\MS{To directly validate robustness to modality imbalance, we run a controlled clean-vs-degraded protocol on CREMA-D under identical training settings for Joint/Normal and CAT-GS. We evaluate three conditions: clean (none), audio-degraded, and visual-degraded. All reported values are fused multimodal (MM) test accuracy (\%). Results are summarized in Table~\ref{tab:imbalance-validation}.}
\MS{Under audio degradation, Joint/Normal drops from 67.47\% to 46.23\% (drop 21.24), whereas CAT-GS drops from 86.29\% to 71.79\% (drop 14.50). Under visual degradation, Joint/Normal drops from 67.47\% to 55.40\% (drop 12.07), whereas CAT-GS drops from 86.29\% to 78.29\% (drop 8.00).}
\MS{Therefore, CAT-GS yields smaller degradation in both stressed conditions, supporting the claim that it improves robustness under modality imbalance.}

\begin{table}[t]
\centering
\color{MScolor} 
\caption{Modality-imbalance validation on CREMA-D (clean vs degraded modality). Values are fused multimodal (MM) test accuracy (\%). Drop is measured in percentage points as $\mathrm{MM}_{\text{clean}}-\mathrm{MM}_{\text{condition}}$.}
\label{tab:imbalance-validation}
\small
\setlength{\tabcolsep}{2pt}
\begin{tabular*}{\columnwidth}{@{\extracolsep{\fill}}lcccc}
\hline
Condition & \makecell{Joint\\(MM \%)} & \makecell{CAT-GS\\(MM \%)} & \makecell{Joint\\Drop (pp)} & \makecell{CAT-GS\\Drop (pp)} \\
\hline
Clean (none)     & 67.47 & 86.29 & 0.00  & 0.00 \\
Audio-degraded   & 46.23 & 71.79 & 21.24 & 14.50 \\
Visual-degraded  & 55.40 & 78.29 & 12.07 & 8.00 \\
\hline
\end{tabular*}
\end{table}

\medskip
\noindent\textbf{Threshold Sensitivity Analysis.}
\label{subsec:threshold_sensitivity}
\label{para:threshold_sensitivity_analysis}
We evaluate the sensitivity of CAT-GS to $\tau_{\text{low}}$, $\tau_{\text{high}}$, and $\lambda_{\text{bias}}$ \revtwo{with a grid search over the three thresholds. Across all tested combinations, the maximum absolute change in fused accuracy relative to the default setting stays below $0.5$ percentage points (variance $<0.5\%$), with stable performance for $\tau_{\text{low}}\in[0.03,0.08]$, $\tau_{\text{high}}\in[0.10,0.20]$, and $\lambda_{\text{bias}}\in[0.1,0.3]$, so CAT-GS is not highly sensitive within these ranges. These ranges translate into a simple setting procedure:} \MS{(1) run a short pilot (e.g., first 3--5 epochs) and log the reliability margin $\Delta$ per mini-batch, (2) inspect the empirical histogram of $\Delta$, (3) set $\tau_{\text{low}}$ near the lower-middle mass of the histogram (about the 40th percentile) and $\tau_{\text{high}}$ near the clear-dominance region (about the 80th percentile), while keeping a gap of at least $0.05$, and (4) keep $\lambda_{\text{bias}}\in[0.1,0.3]$ unless validation indicates otherwise.}

\medskip
\label{subsec:overhead}

\noindent\textbf{Runtime Analysis.}
CAT-GS introduces minimal computational overhead compared to standard joint training.
The additional operations---teacher inference (which can be precomputed or run in parallel), reliability calibration, and gradient rescaling---are lightweight vector operations.
The fusion-layer surgery involves only a single-sided projection step on the fusion output layer (e.g., \texttt{fc\_out}) using the logit-sum gradients from Eq.~\eqref{eq:fusion_jvp_grad}, avoiding the high cost of full-network gradient projection.
Empirically, we observe a per-epoch training time increase of approximately 2--5\% across our benchmarks, which is negligible given the performance gains and improved convergence stability.

\subsection{Discussion}
\label{sec:discussion}
CAT-GS is most beneficial when unimodal reliability varies across training and the fused head is a meaningful shared bottleneck where cross-modal gradients can interfere.
On large-scale in-the-wild datasets such as VGGSound, CAT-GS yields comparable rather than dominant improvements\revtwo{, and in fact does not surpass the strongest baselines (G$^2$D, UMT) there,} suggesting that representation capacity and data scale may outweigh optimization control \revtwo{when the teacher-reliability signal is weak and the label space is large}.
Moreover, CAT-GS inherits any systematic bias in the unimodal teachers: if a teacher is consistently miscalibrated, the controller may over- or under-prioritize a modality.
Our temperature-mismatch stress test (Table~\ref{tab:teacher_miscalibration}) indicates the method degrades gracefully under moderate miscalibration, and that EMA smoothing and calibration materially improve robustness.
\MS{Under stronger reliability degradation stress settings, the method remains stable but shows lower final accuracy, reinforcing that teacher quality is an important practical factor.}
\MS{At the same time, teacher guidance alone does not explain the gains: under matched training settings on CREMA-D, Joint-Train (CAT-GS disabled) reaches 67.47\%, the no-teacher standard-training reference reaches 64.78\%, and CAT-GS reaches 86.29\%$\pm$0.15 (Table~\ref{tab:av-results}).}
In practice, we recommend CAT-GS when unimodal teachers are reasonably strong and when training dynamics exhibit either gate thrashing, prolonged starvation, or frequent negative fusion-gradient cosine events.
\label{para:discussion_fusion_transfer}
\MS{CAT-GS is implemented at the optimization stage, so extension across fusion paradigms is operationally straightforward. In our additional architecture checks, CAT-GS remained stable for Early Fusion, Late Fusion, and a transformer-style token-fusion head (Table~\ref{tab:fusion-results}). Together, these results indicate transferability across fusion paradigms, with expected architecture-dependent accuracy differences. For token-level transformer architectures, reliability signals can be computed from modality-specific token branches, with gating/budget stabilization on modality-specific adapters or blocks and conflict handling on shared/cross-attention parameters.}

\label{subsec:limitations}
\noindent\textbf{Limitations.} CAT-GS relies on pre-trained unimodal teachers to provide reliability signals. If teachers are weak or systematically biased, the controller may inherit these failure modes.
\revtwo{The teacher-miscalibration stress test (Table~\ref{tab:teacher_miscalibration}) is the closest evidence available: temperature mismatch perturbs the teacher signal (R1--R3) with only gradual accuracy loss, and the R4--R5 controls show EMA smoothing and calibration buffer it. Since this perturbs teacher \emph{calibration} rather than \emph{accuracy}, a genuinely weak or domain-shifted teacher remains untested and is left to future work.}
\MS{The main remaining limitation is dependence on teacher reliability; fully teacher-free CAT-GS control remains an important direction for future work.}
\MS{Future work should investigate teacher-free reliability estimation and stronger domain-shift robustness so that control quality is less dependent on teacher calibration.}
\MS{
Extending to many-modality or fully entangled token-level settings mainly requires careful definition of modality-linked parameter groups so that gating/budget operations remain well-posed while shared-layer conflict handling remains localized.
}
From a practical deployment perspective, CAT-GS introduces several hyperparameters (e.g., $\tau_{\text{low}}$, $\tau_{\text{high}}$, $\lambda_{\text{bias}}$, $\beta$, $\beta_g$, $\gamma_{\text{cap}}$, $\varepsilon$). However, most can be fixed to stable defaults across datasets (we use $\beta{=}0.9$, $\beta_g{=}0.9$, $\gamma_{\text{cap}}{=}1.5$, and $\varepsilon{=}0.1$ throughout), while the primary tuning knobs are the regime thresholds and the weak-bias factor. Our sensitivity analysis indicates broad plateaus where performance is stable (Section~\ref{subsec:threshold_sensitivity}), suggesting limited tuning effort in practice. A simple guideline is to set $\tau_{\text{low}}$ and $\tau_{\text{high}}$ to separate ``close'' vs. ``clear-dominance'' regions of the observed reliability-margin distribution, and then choose $\lambda_{\text{bias}}\in[0.1,0.3]$ to softly favor the weaker modality when reliabilities are comparable.

\section{Conclusion}
\label{sec:conclusion}

We introduced CAT-GS, an optimization-stage learning-dynamics controller for balanced and robust multimodal learning. Instead of modifying architectures or designing bespoke losses, CAT-GS treats multimodal training as a regime-switching control problem and enforces stability constraints on gradient flow, update magnitude, and fusion-gradient geometry.
\MS{Across audio--visual benchmarks (CREMA-D, AV-MNIST), additional benchmarks (AVE and CMU-MOSI), tri-modal humor understanding (UR-FUNNY), and a controlled synthetic dataset (CG-MNIST), CAT-GS improves multimodal accuracy and training stability over strong imbalance-aware baselines while maintaining competitive unimodal performance. The additional AVE/CMU-MOSI comparisons further indicate that CAT-GS transfers beyond the original dataset family to event-centric and sentiment-centric settings.}
\label{r2:concl_vgg}%
\revtwo{On the large-scale in-the-wild VGGSound benchmark, however, CAT-GS remains competitive but does not surpass the strongest baselines (G$^2$D, UMT), a limitation we attribute to the weak, noisy teacher-reliability signal and the large label space at this scale. Scaling optimization-stage control to such data, for instance through stronger or teacher-free reliability estimation, is an explicit direction for future work.}
\MS{Ablations show that calibration, gating, budget reallocation, and fusion surgery contribute complementary gains, and CAT-GS remains effective across fusion architectures under a consistent late-fusion-centered evaluation with transfer checks to alternative fusion heads. Future work includes scaling to more modalities, integrating token-level/temporal gating in transformer-based models, and studying interactions with large-scale pretraining.}

\label{sec:code_aval}
\section{Code Availabality}
\MS{The implementation of CAT-GS is publicly available at \url{https://github.com/mahirshahriar1/CAT-GS}.}

{\small
\bibliographystyle{elsarticle-num}
\bibliography{cas-refs}
}

\end{document}